%% file: main.tex
\documentclass[10pt,twocolumn,letterpaper]{article}

\usepackage[T1]{fontenc}
\usepackage[pagenumbers]{cvpr}

\input{preamble}
\definecolor{cvprblue}{rgb}{0.21,0.49,0.74}
\usepackage[pagebackref,breaklinks,colorlinks,allcolors=cvprblue]{hyperref}

\usepackage{url}
\usepackage{graphicx}
\usepackage{booktabs}
\usepackage{float}
\usepackage{enumitem}
\usepackage{multirow}

\def\confName{CVPR}
\def\confYear{2027}

\title{From Diagnosing Alignment to Joint Supervision in Multimodal Draft Training}

\author{%
Haozhe Du\quad
Runze Liu\quad
Jiaming Fan\quad
Yitong Tao\quad
Shuxia Lin\\[3pt]
Chonghan Liu\quad
Daming Cao\quad
Xu Yang
}

\hypersetup{
    pdftitle={From Diagnosing Alignment to Joint Supervision in Multimodal Draft Training},
    pdfauthor={Haozhe Du, Runze Liu, Jiaming Fan, Yitong Tao, Shuxia Lin, Chonghan Liu, Daming Cao, Xu Yang},
    pdfsubject={Multimodal speculative decoding and draft training}
}

\begin{document}
\maketitle

\input{sec/0}

\input{sec/1}

\input{sec/2}
\input{sec/3}

\input{sec/4}
\input{sec/5}
\input{sec/6}
\input{sec/7}

\clearpage
\input{sec/llm_usage}
{
    \small
    \bibliographystyle{ieeenat_fullname}
    \bibliography{main}
}
\clearpage

\appendix

\input{appendix/selection}
\input{appendix/qwen}
\input{appendix/setup}

\end{document}

%% file: preamble.tex
\usepackage{cuted} 

\usepackage{currfile} 

\usepackage{caption} 

%% file: sec/0.tex
\begin{abstract}
Multimodal drafts learn from multimodal (MM) and text-only (TXT)
supervision, yet their effects on alignment and how to combine them
remain unclear.
We diagnose draft--target alignment through paired predictions
with and without visual input, decomposing distributional error
into text-conditioned and visual-response components.
Drafts with similar text-conditioned alignment can still differ
markedly in visual-response risk and acceptance.
Controlled continuation training shows that high-risk groups
outperform random and low-risk groups, and visual-response risk
guides MM training more effectively than aggregate risk.
Building on these diagnostics, we propose \textsc{MNRS-Joint}, which
configures joint supervision from a given draft's component-risk
distributions, using modality-specific mean risks to determine
the MM:TXT mixture.
Across three warm-ups, its distinct mixtures fall in high-performing
regions of the ratio sweeps.
On LLaVA-1.5-7B, \textsc{MNRS-Joint} improves mean accepted length
from 3.35 to 3.53 and mean speedup from $2.04\times$ to
$2.16\times$ over composition-matched random training under greedy
tree decoding. The gains also hold under stochastic chain decoding.
\end{abstract}

%% file: sec/1.tex
\section{Introduction}
\label{sec:introduction}

Autoregressive generation incurs substantial inference latency in
vision-language models~\citep{flamingo2022,blip22023,internvl,cambrian1,qwen2vl}.
Speculative decoding (SD) accelerates generation with a lightweight
draft that proposes tokens for parallel verification by the target
model~\citep{sd-leviathan23a,acceleratinglargelanguagemodel}.
Its effectiveness depends on draft--target
agreement~\citep{distillspec,online_speculative_decoding}, making
alignment a central objective of multimodal draft training.

\begin{figure*}[!t]
    \centering
    \includegraphics[width=0.9\textwidth]
        {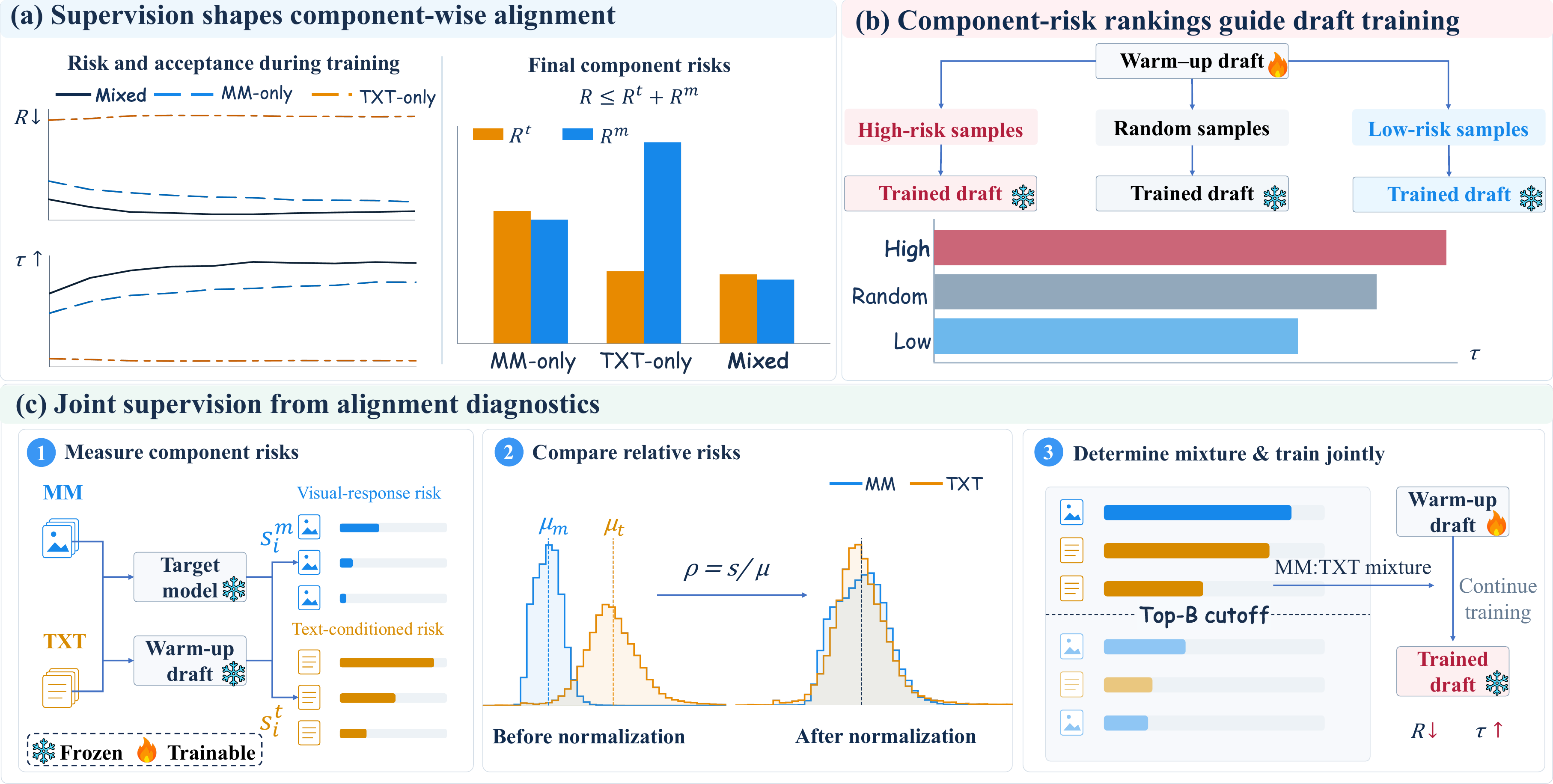}
\caption{
\textbf{Schematic overview of the risk framework.}
(a) Supervision shapes component risks and accepted length.
(b) High-risk groups outperform random and low-risk groups.
(c) Relative component risks determine the MM:TXT mixture.
}
    \label{fig:framework_overview}
   \vspace{-10pt}
\end{figure*}

Multimodal drafts can learn from both multimodal (MM) and text-only
(TXT) supervision. Both sources are also used in vision-language
model training~\citep{MM1,mm15}.
Language-only drafts can accelerate multimodal targets in some
settings~\citep{Gagrani2024}.
MSD combines the two supervision sources through progressive
training~\citep{MSD}, while ViSpec and SpecFLASH incorporate visual
information into draft design~\citep{ViSpec,SpecFLASH}.
These approaches offer different training and design paths.
Further training also calls for considering the draft's current
disagreement with the target.
How can a given draft's alignment state guide joint MM and TXT
supervision to improve decoding performance?

Understanding how each source shapes alignment informs joint
supervision. Aggregate prediction error measures discrepancy
magnitude but does not distinguish text-conditioned disagreement
from differences in visual response.
We decompose the full distributional error into text-conditioned
and visual-response components through paired draft and target
predictions with and without visual input
(Figure~\ref{fig:framework_overview}(a)).
This reveals alignment differences across supervision regimes that
aggregate error obscures:
TXT-only and mixed training reach similar text-conditioned risk,
yet TXT-only retains substantially higher visual-response risk
and produces shorter accepted sequences.
Similar text-conditioned alignment can thus correspond to different
multimodal alignment and speculative decoding performance.

These diagnostics also reveal the training benefits of joint
supervision. Mixed supervision reduces both component risks
relative to MM-only training, even at equal total sample counts.
Controlled continuation further establishes the training value of
component risks: at fixed sample counts and modality composition,
high-risk groups outperform random and low-risk groups.
This ordering also holds when supervision in either modality is
varied separately.
For MM training, visual-response risk provides more effective
guidance than aggregate risk (Figure~\ref{fig:framework_overview}(b)).

Building on these findings, we propose \textsc{MNRS-Joint},
a simple and effective approach to configuring joint supervision
from the warm-up draft's component risks under a limited budget. Using each modality's mean risk as a reference, it preserves
within-modality rankings while adapting the MM:TXT mixture to
the draft's risk distributions (Figure~\ref{fig:framework_overview}(c)).
Ratio sweeps across three warm-up regimes show that the resulting
mixtures differ in composition yet fall in high-performing regions
of their respective sweeps.

With a common EAGLE warm-up~\citep{EAGLE} for
LLaVA-1.5-7B~\citep{LLaVA} and a 34K-example continuation budget,
\textsc{MNRS-Joint} improves mean accepted length from 3.35 to 3.53
and mean speedup from $2.04\times$ to $2.16\times$ over
composition-matched random training under greedy tree decoding.
The gains persist under stochastic chain decoding.
Using exactly the same examples and exposure counts as
\textsc{MNRS-Joint}, sequential TXT-to-MM training yields lower
mean accepted length and speedup under both settings.

Our contributions are twofold.
\begin{itemize}
    \item \textbf{Alignment diagnostics and their training value.}
    We decompose draft--target error into text-conditioned and
    visual-response components, revealing alignment differences
    across supervision regimes that aggregate error obscures.
    Controlled continuation establishes their training value,
    including the advantage of visual-response risk over
    aggregate MM risk.
    \item \textbf{Alignment-guided joint supervision.}
    \textsc{MNRS-Joint} uses these diagnostics to configure MM
    and TXT supervision for a given draft under a shared budget.
    Ratio sweeps across three warm-ups show that its distinct
    mixtures fall in high-performing regions; controlled
    comparisons demonstrate gains under both decoding settings.
\end{itemize}

%% file: sec/2.tex
\section{Related Work}
\label{sec:related_work}

\paragraph{Speculative decoding.}
Draft-and-verify methods include blockwise prediction
\citep{blockwise} and speculative decoding
\citep{SpecDec,sd-leviathan23a}.
Xia et al. survey drafting and verification
strategies~\citep{sd_survey}.
Multi-head and feature-conditioned drafts improve candidate
prediction~\citep{Medusa,hydra,EAGLE,EAGLE3,redrafter}, while
tree-based systems expand candidate coverage and optimize
verification~\citep{specinfer,EAGLE2,sequoia}.
Lookahead Decoding and REST obtain candidates through
Jacobi iteration and retrieval, respectively
\citep{lookahead,rest}.
Other approaches refine block drafts~\citep{draft_refinement},
integrate future-token prediction into the target
\citep{spec_streaming}, or use diffusion-based
parallel drafting~\citep{DFLASH}.
DistillSpec studies distillation objectives and training-data
generation for draft--target agreement~\citep{distillspec}.
LK losses directly target token acceptance~\citep{lk_losses}.
AdaSPEC further improves draft--target alignment through
reference-model-guided token filtering during
distillation~\citep{adaspec}.

\paragraph{Multimodal draft modeling and training.}
Language-only drafts can support multimodal targets~\citep{Gagrani2024}, while ViSpec and SpecFLASH
incorporate visual information into draft
design~\citep{ViSpec,SpecFLASH}.
MSD combines MM and TXT supervision through progressive
training~\citep{MSD}, while TIGER couples text-conditioned
visual routing with verifier-derived accepted-prefix
rewards for draft optimization~\citep{tiger}.
MMSpec provides unified evaluation, distinguishing throughput
from latency gains and examining vision awareness under
batching~\citep{mmspec}.
Supervision composition also matters in multimodal model
training~\citep{MM1,mm15}.
Language-model pretraining methods learn domain mixtures
using proxy-model optimization
\citep{doremi} or performance prediction~\citep{regmix}.
Our approach uses the warm-up draft's component risks to guide
continuation training and determine the MM:TXT supervision mixture,
connecting its current alignment state to the joint use of both sources.

%% file: sec/3.tex
\section{\texorpdfstring{Decomposing Multimodal Draft--Target\\Disagreement}{Decomposing Multimodal Draft--Target Disagreement}}
\label{sec:risk}

We introduce distributional risk as a local diagnostic of multimodal
draft--target alignment, motivated by its relationship with
single-token acceptance under stochastic
verification~\citep{sd-leviathan23a,distillspec}.
We then decompose the distributional error into text-conditioned
and visual-response components.

\subsection{Risk and Single-Token Acceptance}
\label{sec:risk_acceptance}

Let $I$ denote the image input, $T$ the textual prompt, and
$y_{<j}$ the generation prefix before position $j$.
Under the multimodal condition, the target and draft distributions
over the vocabulary $\mathcal V$ are
\begin{equation}
p_j=p(\cdot\mid I,T,y_{<j}), \qquad
q_j=q(\cdot\mid I,T,y_{<j}).
\label{eq:mm_distributions}
\end{equation}
We use unadorned symbols for the multimodal condition and
define the total multimodal risk as
\begin{equation}
R_j=\frac{1}{2}\lVert p_j-q_j\rVert_1.
\label{eq:mm_risk}
\end{equation}

Under standard stochastic speculative sampling
\citep{acceleratinglargelanguagemodel}, a candidate
$z\sim q_j$ is accepted with probability
$\min\{1,p_j(z)/q_j(z)\}$.
Averaging over draft candidates gives the single-token
acceptance probability
\begin{equation}
\begin{aligned}
\alpha_j
&= \mathbb{E}_{z\sim q_j}
\left[
\min\left\{1,\frac{p_j(z)}{q_j(z)}\right\}
\right] \\
&= \sum_{z\in\mathcal V}
\min\left\{p_j(z),q_j(z)\right\} \\
&= 1-R_j.
\end{aligned}
\label{eq:risk_acceptance}
\end{equation}

For the distributions used in stochastic verification, including
sampling transformations, $R_j$ equals the single-token rejection
probability. This identity motivates total variation risk as a
local alignment diagnostic. We compare drafts on shared reference
prefixes and assess the diagnostic's training value through
actual chain and tree decoding.

\subsection{Text-Conditioned and Visual-Response Risks}
\label{sec:risk_decomposition}

We construct a text condition by removing visual tokens while
retaining the textual prompt $T$ and the same generation prefix
$y_{<j}$:
\begin{equation}
p_j^t=p(\cdot\mid T,y_{<j}), \qquad
q_j^t=q(\cdot\mid T,y_{<j}).
\label{eq:text_distributions}
\end{equation}
This paired text condition serves as the diagnostic reference.
Related comparisons with and without visual input assess visual
dependence and mitigate language-prior
bias~\citep{favero2024m3id,liu2024paying}.
VCD contrasts predictions from original and
perturbed images to strengthen visual grounding~\citep{vcd}.

From the draft--target distribution gaps under the multimodal
and text conditions, we define three error terms:
\begin{equation}
\begin{aligned}
\mathbf e_j
&= p_j-q_j,\\
\mathbf e_j^t
&= p_j^t-q_j^t,\\
\mathbf e_j^m
&= (p_j-p_j^t)-(q_j-q_j^t).
\end{aligned}
\label{eq:error_components}
\end{equation}
Here, $\mathbf e_j^t$ measures disagreement under the paired
text condition. The visual-response term $\mathbf e_j^m$ compares
how the two distributions change when the image is added at
the same prefix.
Both components are evaluated on the same example, with
superscripts $t$ and $m$ denoting text-conditioned and
visual-response disagreement, respectively.
The unadorned $\mathbf e_j$ denotes the total multimodal error.
By definition, these three error terms satisfy
\begin{equation}
\mathbf e_j=\mathbf e_j^t+\mathbf e_j^m.
\label{eq:error_decomposition}
\end{equation}

We define the corresponding text-conditioned and visual-response
risks as
\begin{equation}
R_j^t=\frac{1}{2}\lVert\mathbf e_j^t\rVert_1, \qquad
R_j^m=\frac{1}{2}\lVert\mathbf e_j^m\rVert_1.
\label{eq:component_risks}
\end{equation}
The text-conditioned risk $R_j^t$ is the total variation
distance between the target and draft distributions
under the text condition. Both $R_j$ and $R_j^t$
lie in $[0,1]$.
The visual-response risk $R_j^m$ measures the half-$\ell_1$
discrepancy between two signed changes in prediction.
Since $\mathbf e_j^m=\mathbf e_j-\mathbf e_j^t$,
the triangle inequality gives
$0\leq R_j^m\leq R_j+R_j^t\leq 2$.

Applying the triangle inequality to
Equation~(\ref{eq:error_decomposition}) gives
\begin{equation}
R_j \le R_j^t + R_j^m.
\label{eq:risk_upper_bound}
\end{equation}
The two component risks bound the total multimodal mismatch.
The error vectors can partially cancel, so a small total risk
can coexist with larger component risks.
The components thus retain information about the form of
disagreement. We next use them to examine how MM and TXT
supervision shape alignment.

%% file: sec/4.tex
\begin{figure*}[t]
    \centering
    \includegraphics[width=0.9\textwidth]
        {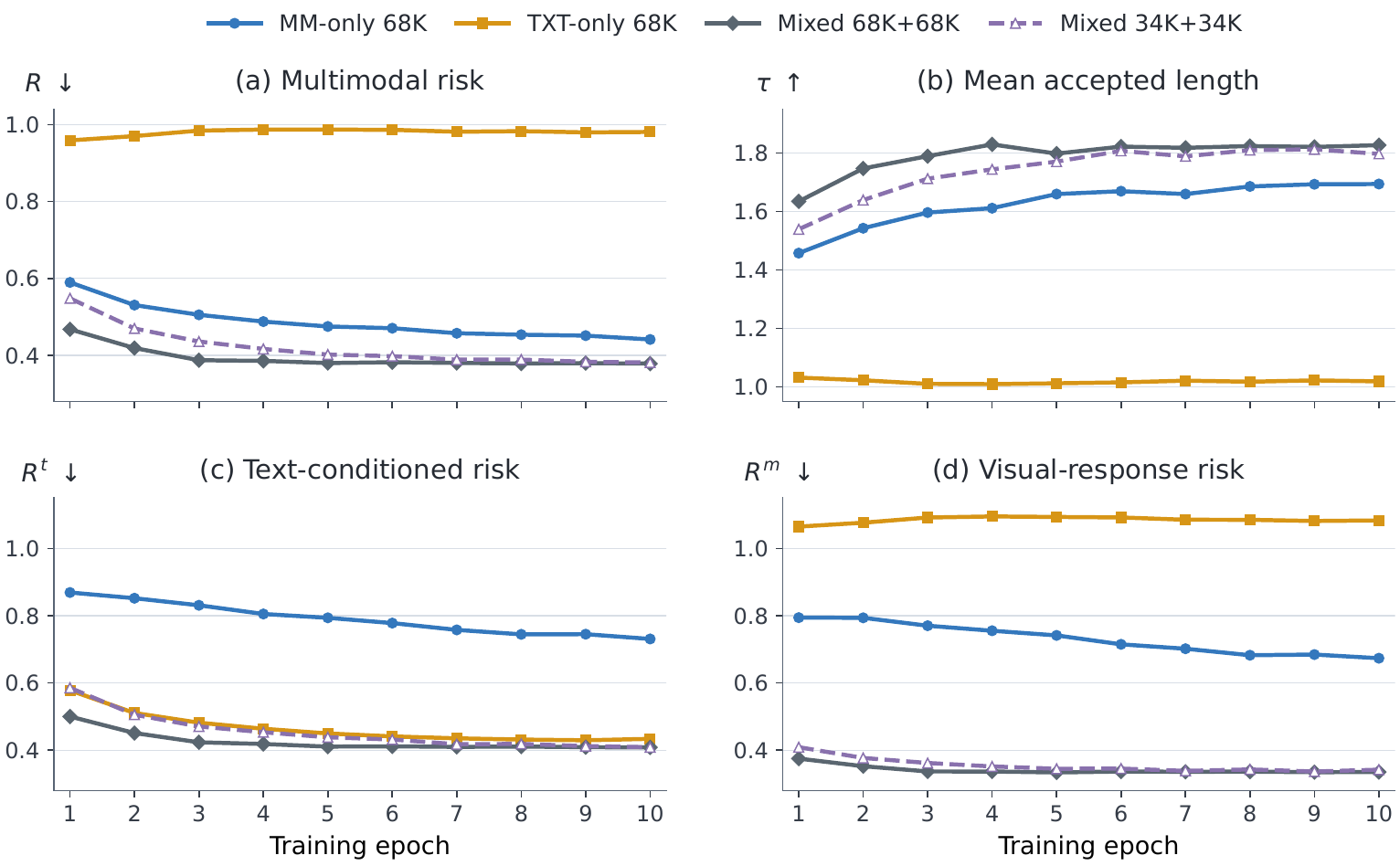}
    \caption{
        \textbf{Risk and accepted length under different
        supervision regimes at $T=1$ with chain decoding.}
        Solid lines show the three primary regimes;
        the dashed line shows the Mixed 34K+34K
        equal-sample-count control.
        Metrics are macro-averaged across five benchmarks.
        Panels (c,d) share the same vertical scale.
    }
    \label{fig:supervision_results}
    \vspace{-10pt}
\end{figure*}

\section{How Supervision Shapes Draft--Target Alignment}
\label{sec:supervision_analysis}

We first examine how MM and TXT supervision affect the two
components of draft--target disagreement.
In particular, we ask whether similar text-conditioned alignment
can coexist with different visual-response risks, and whether
combining MM and TXT supervision improves both components
of draft--target alignment.

\subsection{Supervision Regimes and Evaluation}
\label{sec:supervision_setup}

We use LLaVA-1.5-7B~\citep{LLaVA} as the target model
and train EAGLE drafts~\citep{EAGLE}.
MM and TXT examples come from LLaVA-Instruct-150K
\citep{Visual_instruction_tuning} and ShareGPT, respectively.
We use the same draft initialization and optimizer settings
for all supervision regimes.

We compare \textbf{MM-only} and \textbf{TXT-only}, each using
68K examples, with \textbf{Mixed} supervision in two configurations.
Mixed 68K+68K combines the complete MM and TXT sets used by
the single-modality runs.
Mixed 34K+34K randomly draws 34K examples from each source,
matching the single-modality runs in total sample count.
All runs train for ten epochs, each covering the corresponding
dataset. Mixed 34K+34K therefore matches the single-modality
runs in both total sample count and sample exposure.

We evaluate each checkpoint on ChartQA~\citep{ChartQA},
MM-Vet~\citep{MM-Vet}, ScienceQA~\citep{ScienceQA},
TextVQA~\citep{TextVQA}, and NoCaps~\citep{NoCaps}.
Figure~\ref{fig:supervision_results} reports total risk $R$,
text-conditioned risk $R^t$, visual-response risk $R^m$,
and accepted length $\tau$ under temperature-1 stochastic
chain decoding.
Within each decoding setting, all checkpoints are scored on the
same fixed reference responses and token prefixes. We recompute
target features on these responses and feed them to each draft
for offline risk computation. Accepted lengths are measured
through actual speculative decoding.
Throughput speedup (Sp.) is measured relative to target-only
decoding. All reported means are unweighted arithmetic averages
of the five benchmark-level values, including speedup ratios.
For the continuation experiments, each trained checkpoint is
evaluated under both $T=0$ greedy tree decoding and $T=1$
stochastic chain decoding. Each chain proposes up to four draft
tokens per verification cycle.
Bold marks the best displayed value among continuation
methods within each comparison group and decoding setting,
including ties.
Further evaluation details are provided in Appendix~\ref{app:inference}.

\begin{table*}[!t]
    \centering
    \caption{
    \textbf{Combined within-modality risk grouping.}
    All runs start from the same warm-up and use 17K MM and 17K TXT examples.
    }
    \label{tab:combined_risk_grouping}

    \begingroup
    \footnotesize
    \setlength{\tabcolsep}{2.5pt}
    \renewcommand{\arraystretch}{1.08}

    \resizebox{\textwidth}{!}{%
    \begin{tabular}{@{}l*{15}{c}@{}}
        \toprule[1pt]
        \multirow{2}{*}{Method}
        & \multicolumn{2}{c}{ChartQA}
        & \multicolumn{2}{c}{MM-Vet}
        & \multicolumn{2}{c}{ScienceQA}
        & \multicolumn{2}{c}{TextVQA}
        & \multicolumn{2}{c}{NoCaps}
        & \multicolumn{2}{c}{Mean}
        & \multicolumn{3}{c}{Risk} \\

        \cmidrule(lr){2-3}
        \cmidrule(lr){4-5}
        \cmidrule(lr){6-7}
        \cmidrule(lr){8-9}
        \cmidrule(lr){10-11}
        \cmidrule(lr){12-13}
        \cmidrule(lr){14-16}

        & $\tau$ & Sp.
        & $\tau$ & Sp.
        & $\tau$ & Sp.
        & $\tau$ & Sp.
        & $\tau$ & Sp.
        & $\tau$ & Sp.
        & $R \downarrow$
        & $R^t \downarrow$
        & $R^m \downarrow$ \\
        \specialrule{0.75pt}{3pt}{2pt}

        \multicolumn{16}{c}{\textbf{Temperature=0}} \\
        \midrule[0.3pt]
        Warm-up
        & 2.63 & 1.59$\times$
        & 2.87 & 1.65$\times$
        & 2.68 & 1.61$\times$
        & 2.88 & 1.64$\times$
        & 2.82 & 1.77$\times$
        & 2.78 & 1.65$\times$
        & 0.42 & 0.46 & 0.38 \\

        \midrule[0.3pt]

        \textsc{High}
        & \textbf{3.34} & \textbf{2.07$\times$}
        & \textbf{3.50} & \textbf{2.13$\times$}
        & \textbf{3.52} & \textbf{2.11$\times$}
        & \textbf{3.45} & \textbf{2.20$\times$}
        & \textbf{3.42} & \textbf{2.03$\times$}
        & \textbf{3.45} & \textbf{2.11$\times$}
        & \textbf{0.33}
        & \textbf{0.37}
        & \textbf{0.34} \\

        \textsc{Random}
        & 3.10 & 1.86$\times$
        & 3.23 & 1.90$\times$
        & 3.19 & 1.87$\times$
        & 3.29 & 2.05$\times$
        & 3.24 & 1.93$\times$
        & 3.21 & 1.92$\times$
        & 0.36 & 0.40 & 0.35 \\

        \textsc{Low}
        & 2.81 & 1.68$\times$
        & 2.97 & 1.72$\times$
        & 2.98 & 1.77$\times$
        & 3.02 & 1.76$\times$
        & 2.92 & 1.80$\times$
        & 2.94 & 1.75$\times$
        & 0.40 & 0.45 & 0.37 \\

        \specialrule{0.75pt}{3pt}{2pt}
        \multicolumn{16}{c}{\textbf{Temperature=1}} \\
        \midrule[0.3pt]
        Warm-up
        & 1.64 & 1.23$\times$ & 1.69 & 1.28$\times$ & 1.67 & 1.17$\times$ & 1.61 & 1.22$\times$ & 1.56 & 1.09$\times$ & 1.63 & 1.20$\times$
        & 0.47 & 0.50 & 0.37 \\
        \midrule[0.3pt]

        \textsc{High}
        & \textbf{1.84} & \textbf{1.53$\times$} & \textbf{1.87} & \textbf{1.62$\times$} & \textbf{1.96} & \textbf{1.55$\times$} & \textbf{1.77} & \textbf{1.61$\times$} & \textbf{1.62} & \textbf{1.19$\times$} & \textbf{1.81} & \textbf{1.50$\times$}
        & \textbf{0.38} & \textbf{0.41} & \textbf{0.34} \\

        \textsc{Random}
        & 1.82 & 1.50$\times$ & 1.82 & 1.50$\times$ & 1.88 & 1.48$\times$ & 1.70 & 1.50$\times$ & 1.58 & 1.14$\times$ & 1.76 & 1.42$\times$
        & 0.41 & 0.44 & 0.35 \\

        \textsc{Low}
        & 1.69 & 1.27$\times$ & 1.73 & 1.33$\times$ & 1.71 & 1.21$\times$ & 1.69 & 1.45$\times$ & 1.53 & 0.99$\times$ & 1.67 & 1.25$\times$
        & 0.45 & 0.48 & 0.37 \\

        \bottomrule[0.8pt]
    \end{tabular}%
    }
    \endgroup
\end{table*}

\subsection{Component-Wise Alignment across Regimes}
\label{sec:supervision_validation}

\paragraph{Overall alignment and acceptance.}
Lower total risk generally accompanies longer accepted sequences
across the evaluated regimes
(Figure~\ref{fig:supervision_results}(a,b)).
At epoch 10, Mixed 68K+68K reaches $R=0.38$ and $\tau=1.83$,
compared with $R=0.44$ and $\tau=1.69$ for MM-only.
TXT-only training retains high total risk and short accepted
lengths. Mixed training brings the draft closer to the target
and yields longer accepted sequences throughout the evaluated
trajectory.

\paragraph{Component risks.}
TXT-only approaches Mixed 68K+68K in text-conditioned risk
but retains substantially greater visual-response risk
(Figure~\ref{fig:supervision_results}(c,d)).
The visual-response component distinguishes drafts that appear
similarly aligned under the text condition, revealing a discrepancy
that accompanies the acceptance gap.
Mixed 68K+68K also improves both components relative to MM-only.
At epoch 10, $R^t$ decreases from 0.73 to 0.41 and $R^m$
from 0.67 to 0.33.
The benefit of combining MM and TXT supervision therefore spans
both text-conditioned and visual-response alignment.

\paragraph{Equal-sample-count control.}
Mixed 34K+34K reaches $\tau=1.80$ and $R=0.38$ at epoch 10,
with component risks $R^t=0.41$ and $R^m=0.34$.
It improves both component risks and acceptance over MM-only
at the same total sample count and exposure, and its accepted
length is close to that of Mixed 68K+68K.
Combining supervision sources thus improves alignment even under
a fixed sample budget.

Together, these comparisons show that MM and TXT supervision
leave distinct patterns of disagreement. The component risks
make these patterns available as signals for further draft training.

%% file: sec/5.tex
\section{Training Implications of Component Risks}
\label{sec:sample_level}

We test whether the component diagnostics from
Section~\ref{sec:supervision_analysis} guide continuation training.
We also compare visual-response and aggregate risk as MM training signals.

\subsection{\texorpdfstring{Risk-Based Grouping and Controlled\\Continuation}{Risk-Based Grouping and Controlled Continuation}}
\label{sec:risk_grouping}

We use the draft obtained after one epoch of Mixed 68K+68K
training as the common warm-up.
This checkpoint both scores the candidates and initializes
all continuation branches.
For each example $x_i$, we average component risks over its
$L_i$ valid response positions:
\begin{equation}
s_i^t=\frac{1}{L_i}\sum_{j=1}^{L_i}R_{i,j}^t,
\qquad
s_i^m=\frac{1}{L_i}\sum_{j=1}^{L_i}R_{i,j}^m.
\label{eq:sample_risk}
\end{equation}
TXT examples are ranked by text-conditioned risk $s_i^t$,
and MM examples by visual-response risk $s_i^m$.
The two rankings order candidates separately within each modality.

Each branch uses 17K MM and 17K TXT examples and trains
for ten additional epochs.
\textsc{High} uses the highest-ranked examples from each
modality, \textsc{Low} uses the lowest-ranked examples,
and \textsc{Random} samples independently from both pools.
Evaluation follows Section~\ref{sec:supervision_setup},
including both decoding settings and benchmark-level averaging.

\subsection{Training Outcomes across Risk Groups}
\label{sec:combined_risk_grouping}

\paragraph{Combined grouping.}
We first test whether the warm-up draft's component risks identify
useful continuation supervision at a fixed budget and composition.
Table~\ref{tab:combined_risk_grouping} shows that
\textsc{High} outperforms \textsc{Random} and \textsc{Low}
under these matched conditions.
Under greedy tree decoding (Table~\ref{tab:combined_risk_grouping}, $T=0$),
their mean accepted lengths are 3.45, 3.21, and 2.94,
respectively, with corresponding mean speedups of
$2.11\times$, $1.92\times$, and $1.75\times$.
The accepted-length ordering holds on all five benchmarks.
The ordering in both mean accepted length and speedup also
holds under stochastic chain decoding
(Table~\ref{tab:combined_risk_grouping}, $T=1$).
After continuation, \textsc{High} yields the lowest mean
total and component risks in both settings, alongside better
drafting performance. At fixed modality counts, supervision
targeting larger warm-up disagreements thus improves both
alignment and acceptance.

\paragraph{Isolating MM and TXT effects.}
We then test the contribution of each risk signal by varying
one modality while fixing the other modality's
\textsc{Random} subset.
The MM branches vary $s_i^m$-ranked supervision with TXT
fixed, and the TXT branches vary $s_i^t$-ranked supervision
with MM fixed.
All remaining settings are unchanged.

\begin{table}[!tb]
    \centering
    \caption{
    \textbf{Controlled component-risk grouping.}
    Only the indicated modality varies; the other uses
    \textsc{Random}.
    }
    \label{tab:single_risk_ablation}

    \begingroup
    \footnotesize
    \setlength{\tabcolsep}{3pt}
    \renewcommand{\arraystretch}{1.08}

    \begin{tabular*}{\columnwidth}{@{\extracolsep{\fill}}lccccc@{}}
        \toprule[1pt]
        Method
        & $\tau \uparrow$
        & Sp.$\uparrow$
        & $R \downarrow$
        & $R^t \downarrow$
        & $R^m \downarrow$ \\
        \specialrule{0.75pt}{3pt}{2pt}

        \multicolumn{6}{c}{\textbf{Temperature=0}} \\
        \midrule[0.3pt]
        MM-\textsc{High}
        & \textbf{3.37}
        & \textbf{2.06$\times$}
        & \textbf{0.34}
        & \textbf{0.38}
        & \textbf{0.34} \\

        MM-\textsc{Random}
        & 3.21
        & 1.92$\times$
        & 0.36 & 0.40 & 0.35 \\

        MM-\textsc{Low}
        & 3.02
        & 1.80$\times$
        & 0.39 & 0.43 & 0.36 \\

        \midrule

        TXT-\textsc{High}
        & \textbf{3.35}
        & \textbf{2.02$\times$}
        & \textbf{0.35}
        & \textbf{0.38}
        & \textbf{0.34} \\

        TXT-\textsc{Random}
        & 3.21
        & 1.92$\times$
        & 0.36 & 0.40 & 0.35 \\

        TXT-\textsc{Low}
        & 3.16
        & 1.88$\times$
        & 0.37 & 0.41 & 0.36 \\

        \specialrule{0.75pt}{3pt}{2pt}
        \multicolumn{6}{c}{\textbf{Temperature=1}} \\
        \midrule[0.3pt]
        MM-\textsc{High} & \textbf{1.79} & \textbf{1.45$\times$} & \textbf{0.39} & \textbf{0.43} & \textbf{0.34} \\
        MM-\textsc{Random} & 1.76 & 1.42$\times$ & 0.41 & 0.44 & 0.35 \\
        MM-\textsc{Low} & 1.69 & 1.28$\times$ & 0.43 & 0.47 & 0.36 \\
        \midrule
        TXT-\textsc{High} & \textbf{1.80} & \textbf{1.48$\times$} & \textbf{0.39} & \textbf{0.42} & \textbf{0.35} \\
        TXT-\textsc{Random} & 1.76 & 1.42$\times$ & 0.41 & 0.44 & \textbf{0.35} \\
        TXT-\textsc{Low} & 1.72 & 1.33$\times$ & 0.42 & 0.45 & 0.36 \\
        \bottomrule[0.8pt]
    \end{tabular*}
    \endgroup
\end{table}

Table~\ref{tab:single_risk_ablation} shows that the
High--Random--Low ordering in mean accepted length
persists when either modality is varied separately.
Under greedy tree decoding, the high-risk MM and TXT
branches reach $\tau=3.37$
and $3.35$, respectively, compared with $3.21$ for the
shared random baseline.
Their mean speedups are $2.06\times$ and $2.02\times$,
compared with $1.92\times$ for random continuation.
The same ordering in both mean metrics also holds under
stochastic chain decoding
(Table~\ref{tab:single_risk_ablation}, $T=1$).
In both settings, \textsc{High} also achieves the lowest
post-training total and component risks within each comparison,
including ties at the displayed precision.
With the other supervision source fixed, each component risk
provides useful guidance for training within its modality.

\subsection{Visual-Response Risk versus Aggregate Risk}
\label{sec:component_vs_aggregate}

We next ask whether distinguishing the visual-response component
improves MM training beyond using overall disagreement.
For MM examples, we compare ranking by $s_i^m$ with ranking
by the aggregate-risk score
\begin{equation}
s_i=\frac{1}{L_i}\sum_{j=1}^{L_i}R_{i,j}.
\label{eq:sample_aggregate_risk}
\end{equation}
Both branches use 17K MM examples and the same 17K
randomly sampled TXT subset, warm-up checkpoint, and
training settings.
Only the MM ranking signal changes.
The visual-response branch is the same MM-\textsc{High}
run reported in Table~\ref{tab:single_risk_ablation}.

\begin{table}[H]
    \centering
    \caption{
    \textbf{Visual-response versus aggregate risk.}
    Both strategies use 17K MM examples and the same
    17K random TXT subset.
    }
    \label{tab:component_vs_aggregate}

    \begingroup
    \footnotesize
    \setlength{\tabcolsep}{3pt}
    \renewcommand{\arraystretch}{1.08}

    \begin{tabular*}{\columnwidth}{@{\extracolsep{\fill}}lccccc@{}}
        \toprule[1pt]
        MM ranking signal
        & $\tau \uparrow$
        & Sp.$\uparrow$
        & $R \downarrow$
        & $R^t \downarrow$
        & $R^m \downarrow$ \\
        \specialrule{0.75pt}{3pt}{2pt}

        \multicolumn{6}{c}{\textbf{Temperature=0}} \\
        \midrule[0.3pt]
        Aggregate risk $s_i$
        & 3.28
        & 1.97$\times$
        & \textbf{0.34}
        & \textbf{0.38}
        & \textbf{0.34} \\

        Visual-response $s_i^m$
        & \textbf{3.37}
        & \textbf{2.06$\times$}
        & \textbf{0.34}
        & \textbf{0.38}
        & \textbf{0.34} \\

        \specialrule{0.75pt}{3pt}{2pt}
        \multicolumn{6}{c}{\textbf{Temperature=1}} \\
        \midrule[0.3pt]
        Aggregate risk $s_i$ & 1.73 & 1.37$\times$ & 0.40 & \textbf{0.43} & \textbf{0.34} \\
        Visual-response $s_i^m$ & \textbf{1.79} & \textbf{1.45$\times$} & \textbf{0.39} & \textbf{0.43} & \textbf{0.34} \\
        \bottomrule[0.8pt]
    \end{tabular*}
    \endgroup
\end{table}

Under greedy tree decoding
(Table~\ref{tab:component_vs_aggregate}, $T=0$),
visual-response ranking raises mean accepted length from
3.28 to 3.37 and speedup from $1.97\times$ to $2.06\times$.
Under stochastic chain decoding, mean accepted length rises
from 1.73 to 1.79 and speedup from $1.37\times$ to $1.45\times$,
while total risk decreases from 0.40 to 0.39.
With the TXT subset, MM budget, and continuation settings fixed,
the gains show that visual-response risk provides more effective
guidance for MM continuation training than aggregate risk.
Additional target-model experiments are provided in
Appendix~\ref{app:qwen_selection}.

Together, these comparisons establish the value of component
diagnostics for guiding supervision within each modality at fixed
sample counts. The next section uses a given draft's risk
distributions to combine MM and TXT supervision, addressing its
remaining alignment gaps to improve decoding performance under
a shared budget.

%% file: sec/6.tex
\begin{table*}[!t]
    \centering
    \caption{
    \textbf{Combining MM and TXT supervision through relative risk.}
    All runs start from the same warm-up and use 34K examples.
    }
    \label{tab:joint_selection}

    \begingroup
    \footnotesize
    \setlength{\tabcolsep}{2.5pt}
    \renewcommand{\arraystretch}{1.16}

    \resizebox{\textwidth}{!}{%
    \begin{tabular}{@{}l*{15}{c}@{}}
        \toprule[1pt]
        \multirow{2}{*}{Method}
        & \multicolumn{2}{c}{ChartQA}
        & \multicolumn{2}{c}{MM-Vet}
        & \multicolumn{2}{c}{ScienceQA}
        & \multicolumn{2}{c}{TextVQA}
        & \multicolumn{2}{c}{NoCaps}
        & \multicolumn{2}{c}{Mean}
        & \multicolumn{3}{c}{Risk} \\

        \cmidrule(lr){2-3}
        \cmidrule(lr){4-5}
        \cmidrule(lr){6-7}
        \cmidrule(lr){8-9}
        \cmidrule(lr){10-11}
        \cmidrule(lr){12-13}
        \cmidrule(lr){14-16}

        & $\tau$ & Sp.
        & $\tau$ & Sp.
        & $\tau$ & Sp.
        & $\tau$ & Sp.
        & $\tau$ & Sp.
        & $\tau$ & Sp.
        & $R \downarrow$
        & $R^t \downarrow$
        & $R^m \downarrow$ \\
        \specialrule{0.75pt}{3pt}{2pt}

        \multicolumn{16}{c}{\textbf{Temperature=0}} \\
        \midrule[0.3pt]
        Warm-up
        & 2.63 & 1.59$\times$
        & 2.87 & 1.65$\times$
        & 2.68 & 1.61$\times$
        & 2.88 & 1.64$\times$
        & 2.82 & 1.77$\times$
        & 2.78 & 1.65$\times$
        & 0.42 & 0.46 & 0.38 \\

        \midrule[0.3pt]

        Raw-Joint
        & 3.06 & 1.84$\times$
        & 3.02 & 1.81$\times$
        & 3.26 & 1.91$\times$
        & 3.01 & 1.76$\times$
        & 2.78 & 1.75$\times$
        & 3.03 & 1.81$\times$
        & 0.39 & 0.41 & 0.37 \\

        Random-Matched
        & 3.30 & 2.04$\times$
        & 3.42 & 2.06$\times$
        & 3.37 & 2.00$\times$
        & 3.33 & 2.11$\times$
        & 3.33 & 1.99$\times$
        & 3.35 & 2.04$\times$
        & 0.34 & 0.38 & \textbf{0.34} \\

        TXT$\rightarrow$MM
        & 2.98 & 1.78$\times$
        & 3.53 & 2.15$\times$
        & 3.20 & 1.88$\times$
        & 3.40 & 2.13$\times$
        & 3.59 & 2.09$\times$
        & 3.34 & 2.01$\times$
        & 0.35 & 0.41 & 0.35 \\

        \midrule

        \textsc{MNRS-Joint}
        & \textbf{3.44} & \textbf{2.15$\times$}
        & \textbf{3.60} & \textbf{2.17$\times$}
        & \textbf{3.54} & \textbf{2.13$\times$}
        & \textbf{3.47} & \textbf{2.23$\times$}
        & \textbf{3.60} & \textbf{2.11$\times$}
        & \textbf{3.53} & \textbf{2.16$\times$}
        & \textbf{0.32}
        & \textbf{0.37}
        & \textbf{0.34} \\

        \specialrule{0.75pt}{3pt}{2pt}
        \multicolumn{16}{c}{\textbf{Temperature=1}} \\
        \midrule[0.3pt]
        Warm-up
        & 1.64 & 1.23$\times$ & 1.69 & 1.28$\times$ & 1.67 & 1.17$\times$ & 1.61 & 1.22$\times$ & 1.56 & 1.09$\times$ & 1.63 & 1.20$\times$
        & 0.47 & 0.50 & 0.37 \\
        \midrule[0.3pt]

        Raw-Joint
        & 1.78 & 1.43$\times$ & 1.74 & 1.39$\times$ & 1.97 & 1.57$\times$ & 1.69 & 1.44$\times$ & 1.43 & 0.74$\times$ & 1.72 & 1.31$\times$
        & 0.43 & 0.44 & 0.37 \\

        Random-Matched
        & 1.73 & 1.31$\times$ & 1.78 & 1.42$\times$ & 1.81 & 1.34$\times$ & 1.69 & 1.44$\times$ & 1.65 & 1.26$\times$ & 1.73 & 1.35$\times$
        & 0.42 & 0.45 & 0.35 \\

        TXT$\rightarrow$MM
        & 1.79 & 1.45$\times$ & \textbf{1.86} & \textbf{1.54$\times$} & 1.85 & 1.41$\times$ & 1.70 & 1.46$\times$ & 1.59 & 1.15$\times$ & 1.76 & 1.40$\times$
        & 0.40 & 0.44 & \textbf{0.34} \\

        \midrule

        \textsc{MNRS-Joint}
        & \textbf{1.88} & \textbf{1.61$\times$} & 1.85 & 1.53$\times$ & \textbf{2.05} & \textbf{1.71$\times$} & \textbf{1.74} & \textbf{1.55$\times$} & \textbf{1.68} & \textbf{1.40$\times$} & \textbf{1.84} & \textbf{1.56$\times$}
        & \textbf{0.38} & \textbf{0.41} & \textbf{0.34} \\

        \bottomrule[0.8pt]
    \end{tabular}%
    }
    \endgroup
\end{table*}

\section{Joint Supervision through Relative Risk}
\label{sec:joint_selection}

\begin{figure}[H]
    \centering
    \includegraphics[width=0.9\columnwidth]
        {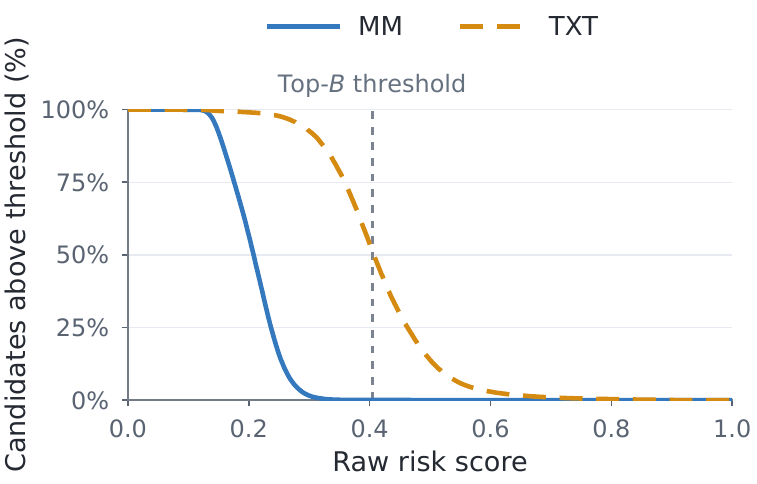}
    \caption{
        \textbf{Raw risk distributions across modalities.}
        Curves show the fraction of candidates in each
        modality exceeding a given score.
        The vertical line marks the joint Top-$B$
        threshold for $B=34\mathrm{K}$.
    }
    \label{fig:raw_risk_distribution}
\end{figure}

Sections~\ref{sec:supervision_analysis} and~\ref{sec:sample_level}
show how MM and TXT supervision shape alignment and how component
risks guide further training. We now combine these sources for a
given warm-up draft under a shared budget, scoring MM and TXT
candidates by visual-response risk $s_i^m$ and text-conditioned risk
$s_i^t$, respectively. However, their raw distributions are markedly
imbalanced (Figure~\ref{fig:raw_risk_distribution}). With the Mixed
warm-up, the means are $\mu_m=0.2082$ and $\mu_t=0.4139$.
Ranking these raw scores retains only 18 MM examples among the
top 34K, allocating almost the entire budget to TXT.

\subsection{From Raw Scores to Relative Risk}
\label{sec:risk_calibration}

We propose \textbf{mean-normalized risk selection with joint pooling
(MNRS-Joint)} to configure joint supervision using relative component
risks (Algorithm~\ref{alg:mnrs_joint}).

Normalization preserves within-modality rankings;
$\rho_i=1$ denotes average risk within a modality.
At $B=34\mathrm{K}$, the Mixed warm-up yields
18,173 MM and 15,827 TXT examples.
Although both normalized pools have mean one, their counts above
the joint cutoff can differ. The cutoff determines where each
modality's ranked list is truncated, specifying both the retained
samples and the budget allocation. \textsc{MNRS-Joint} thus extends
component-risk training guidance to joint supervision, using the
current draft's risk distributions for both decisions.
All scores and means come from the same warm-up checkpoint,
fixing the mixture before continuation training. The ratio sweeps
then assess the resulting allocation.

\newfloat{mnrsalgorithm}{tbp}{loa}
\floatname{mnrsalgorithm}{Algorithm}
\begin{mnrsalgorithm}[H]
    \centering
    \caption{\textbf{MNRS-Joint for joint supervision.}}
    \label{alg:mnrs_joint}
    \small
    \setlength{\tabcolsep}{3pt}
    \renewcommand{\arraystretch}{1.08}
    \begin{tabular}{@{}r p{\dimexpr\columnwidth-18pt\relax}@{}}
        \toprule
        & \textbf{Input:} Warm-up draft, MM/TXT pools, budget $B$. \\
        1 & Score MM by $s_i^m$ and TXT by $s_i^t$. \\
        2 & Compute pool means $\mu_m$ and $\mu_t$. \\
        3 & Set $\rho_i=s_i^m/\mu_m$ for MM and
            $\rho_i=s_i^t/\mu_t$ for TXT. \\
        4 & Pool candidates; retain the top $B$ by $\rho_i$. \\
        & \textbf{Output:} Joint set and MM:TXT sample counts. \\
        \bottomrule
    \end{tabular}
\end{mnrsalgorithm}

\begin{figure*}[!t]
    \centering
    \includegraphics[width=0.96\textwidth]
        {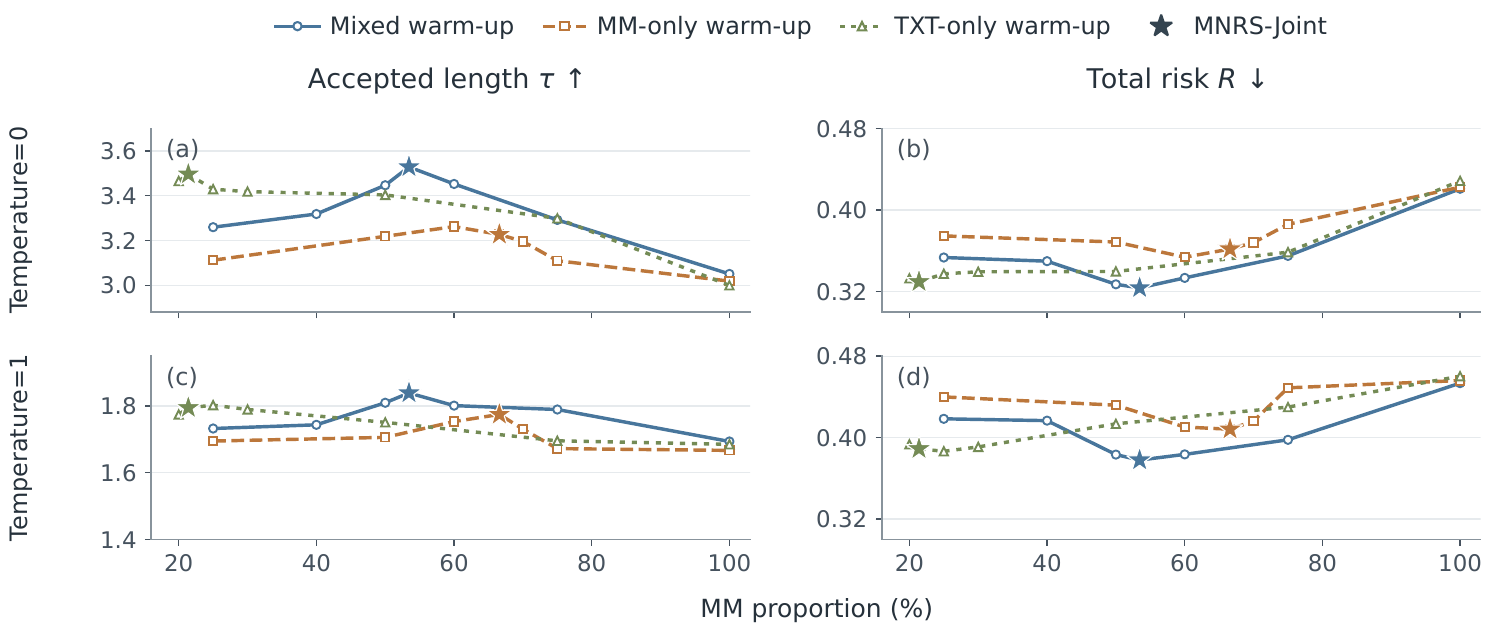}
    \caption{
        \textbf{Modality composition across warm-up regimes.}
        Mean accepted length (left) and total risk (right),
        macro-averaged across five benchmarks, under $T=0$ tree (top)
        and $T=1$ chain (bottom) decoding.
        Stars mark MNRS-Joint ratios; lines connect evaluated ratios.
    }
    \label{fig:modality_composition}
\end{figure*}

\subsection{Joint versus Sequential Continuation}
\label{sec:joint_evaluation}

All methods in Table~\ref{tab:joint_selection} share the same
warm-up and optimization settings (Section~\ref{sec:risk_grouping}),
using 34K examples with ten exposures per example.
Raw-Joint pools unnormalized component scores.
Random-Matched uses the MM:TXT counts induced by
\textsc{MNRS-Joint} and samples randomly within each modality.
TXT$\rightarrow$MM adopts the text-then-multimodal training
order used in ViSpec~\citep{ViSpec}, using exactly the same
examples as \textsc{MNRS-Joint} with matched exposure counts.

\paragraph{Risk guidance at a fixed composition.}
Under greedy tree decoding, \textsc{MNRS-Joint} achieves mean
accepted length 3.53 and speedup $2.16\times$, compared with
3.35 and $2.04\times$ for Random-Matched.
Under stochastic chain decoding, the corresponding values
increase from 1.73 and $1.35\times$ to 1.84 and $1.56\times$.
With modality counts held fixed, these gains demonstrate the
training value of component risks within each modality.

\paragraph{Joint versus sequential supervision.}
TXT$\rightarrow$MM reaches mean accepted length 3.34 and speedup
$2.01\times$ under greedy tree decoding.
\textsc{MNRS-Joint} achieves higher mean accepted length and speedup
under both decoding settings, favoring joint supervision over
sequential training with identical examples and exposure counts.

\textsc{MNRS-Joint} also yields the lowest mean total risk:
$R=0.32$ under greedy tree decoding and $R=0.38$ under stochastic
chain decoding. It leads accepted length and speedup on all five
benchmarks under greedy tree decoding, while TXT$\rightarrow$MM
leads on MM-Vet under stochastic chain decoding.
Raw-Joint reaches 3.03 and $1.81\times$ under greedy tree decoding.
Appendix~\ref{app:selection_budget} reports additional
continuation-training results across different budgets.

\subsection{Evaluating Mean-Normalized Mixtures}
\label{sec:modality_composition}

To evaluate whether mean normalization yields effective supervision
mixtures for different warm-up drafts, we examine the resulting
allocations and their positions on the performance curves
(Figure~\ref{fig:modality_composition}).
The three warm-ups are the checkpoints after one epoch of
Mixed, MM-only, and TXT-only training.
For each warm-up, we vary the MM:TXT ratio at a fixed 34K budget,
keeping within-modality rankings and training settings unchanged.
The resulting checkpoints are evaluated under both decoding settings.

\paragraph{Draft-dependent allocations.}
MNRS-Joint assigns MM proportions of 53.45\%, 66.58\%, and
21.40\% for the Mixed, MM-only, and TXT-only warm-ups,
respectively. These allocations track the shifts in favorable
regions: the TXT-only warm-up favors a TXT-heavy mixture,
whereas the MM-only warm-up favors a larger MM share.

\paragraph{Performance of the resulting mixtures.}
Each MNRS-Joint mixture falls in a high-performing region.
Under greedy tree decoding, it attains the highest
observed accepted length with Mixed and TXT-only warm-ups
and remains within 1.1\% of the best scanned value under MM-only
warm-up. Under stochastic chain decoding, it attains the highest
observed accepted length under Mixed and MM-only warm-ups
and remains within 0.43\% of the best scanned value under TXT-only
warm-up. The corresponding total risks are also at or near the
observed minima in both settings.
Across all three warm-ups, moving from the favorable mixed region
to MM-only continuation lowers accepted length and raises total
risk under both decoding settings. With within-modality rankings
held fixed, these changes show how the supervision balance affects
both alignment and acceptance.

The sweeps validate allocations computed before continuation
training. Together with the fixed-composition controls in
Table~\ref{tab:joint_selection}, they support using modality means
to translate component diagnostics into effective joint supervision
for a given draft.

%% file: sec/7.tex
\section{Conclusion}
\label{sec:conclusion}

We study the roles of MM and TXT supervision in multimodal draft
training through component-wise alignment diagnostics.
Text-conditioned and visual-response components reveal alignment
differences across supervision regimes, and controlled continuation
experiments demonstrate that component risks guide further training.
Visual-response risk guides MM continuation training more effectively
than aggregate risk.
Building on these diagnostics, \textsc{MNRS-Joint} uses
modality-specific mean risks as references to configure joint
supervision for a given draft's remaining alignment gaps.
It determines distinct MM:TXT mixtures across three warm-up
regimes, and ratio sweeps confirm that each lies in a
high-performing region.
The shifts in favorable mixtures highlight the value of adapting
supervision to the current draft's alignment state, using diagnostics
computed before continuation training.
The gains from joint supervision hold under both greedy tree
and stochastic chain decoding.
Together, these results connect alignment diagnostics to improved
drafting performance through both supervision within each modality
and the allocation of training across modalities.

%% file: sec/llm_usage.tex
\section*{LLM Usage Statement}

We used ChatGPT and OpenAI Codex to assist with manuscript
translation, text revision and organization, literature
retrieval, and presentation of results, as well as figure
preparation and LaTeX formatting.
The authors take full responsibility for the final manuscript,
figures, code, experimental results, citations, and scientific
conclusions.

%% file: appendix/selection.tex
\twocolumn[{
\section{Supplementary Experiments and Analyses}
\label{app:selection}

\subsection{Continuation Supervision across Training Budgets}
\label{app:selection_budget}

\begin{minipage}{\textwidth}
\centering
\captionof{table}{\textbf{Continuation supervision across training budgets.}
Within each budget, the two methods use identical modality counts.}
\label{tab:selection_budget}
\footnotesize
\setlength{\tabcolsep}{2pt}
\renewcommand{\arraystretch}{1.04}
\resizebox{0.99\textwidth}{!}{%
\begin{tabular}{@{}l@{\hspace{5pt}}l*{6}{@{\hspace{4pt}}c@{\hspace{2pt}}c}@{\hspace{5pt}}ccc@{}}
\toprule
 & & \multicolumn{2}{c}{ChartQA} & \multicolumn{2}{c}{MM-Vet} & \multicolumn{2}{c}{ScienceQA} & \multicolumn{2}{c}{TextVQA} & \multicolumn{2}{c}{NoCaps} & \multicolumn{2}{c}{Mean} & \multicolumn{3}{c}{Risk} \\
\cmidrule(lr){3-4}\cmidrule(lr){5-6}\cmidrule(lr){7-8}\cmidrule(lr){9-10}\cmidrule(lr){11-12}\cmidrule(lr){13-14}\cmidrule(lr){15-17}
Budget & Method & $\tau$ & Sp. & $\tau$ & Sp. & $\tau$ & Sp. & $\tau$ & Sp. & $\tau$ & Sp. & $\tau$ & Sp. & $R\downarrow$ & $R^t\downarrow$ & $R^m\downarrow$ \\
\specialrule{0.75pt}{3pt}{2pt}
\multicolumn{17}{c}{\textbf{Temperature=0}} \\
\midrule[0.3pt]
 & Warm-up & 2.63 & 1.59$\times$ & 2.87 & 1.65$\times$ & 2.68 & 1.61$\times$ & 2.88 & 1.64$\times$ & 2.82 & 1.77$\times$ & 2.78 & 1.65$\times$ & 0.42 & 0.46 & 0.38 \\
\midrule[0.3pt]
\multirow{2}{*}{17K} & \textsc{MNRS-Joint} & \textbf{2.85} & \textbf{1.72$\times$} & \textbf{3.19} & \textbf{1.84$\times$} & \textbf{2.94} & \textbf{1.80$\times$} & \textbf{3.08} & \textbf{1.84$\times$} & \textbf{3.33} & \textbf{1.99$\times$} & \textbf{3.08} & \textbf{1.84$\times$} & \textbf{0.38} & \textbf{0.42} & \textbf{0.36} \\
 & Random-Matched & \textbf{2.85} & 1.67$\times$ & 3.00 & 1.72$\times$ & 2.80 & 1.66$\times$ & 2.96 & 1.75$\times$ & 2.92 & 1.81$\times$ & 2.90 & 1.72$\times$ & 0.42 & 0.45 & 0.38 \\
\addlinespace[3pt]
\multirow{2}{*}{34K} & \textsc{MNRS-Joint} & \textbf{3.44} & \textbf{2.15$\times$} & \textbf{3.60} & \textbf{2.17$\times$} & \textbf{3.54} & \textbf{2.13$\times$} & \textbf{3.47} & \textbf{2.23$\times$} & \textbf{3.60} & \textbf{2.11$\times$} & \textbf{3.53} & \textbf{2.16$\times$} & \textbf{0.32} & \textbf{0.37} & \textbf{0.34} \\
 & Random-Matched & 3.30 & 2.04$\times$ & 3.42 & 2.06$\times$ & 3.37 & 2.00$\times$ & 3.33 & 2.11$\times$ & 3.33 & 1.99$\times$ & 3.35 & 2.04$\times$ & 0.34 & 0.38 & \textbf{0.34} \\
\addlinespace[3pt]
\multirow{2}{*}{68K} & \textsc{MNRS-Joint} & \textbf{3.61} & \textbf{2.24$\times$} & \textbf{3.68} & \textbf{2.19$\times$} & \textbf{4.02} & \textbf{2.51$\times$} & \textbf{3.53} & \textbf{2.32$\times$} & \textbf{3.47} & \textbf{2.02$\times$} & \textbf{3.66} & \textbf{2.26$\times$} & \textbf{0.32} & \textbf{0.35} & \textbf{0.32} \\
 & Random-Matched & 3.42 & 2.14$\times$ & 3.56 & 2.12$\times$ & 3.74 & 2.28$\times$ & 3.41 & 2.12$\times$ & 3.27 & 1.98$\times$ & 3.48 & 2.13$\times$ & 0.34 & 0.37 & 0.33 \\
\specialrule{0.75pt}{3pt}{2pt}
\multicolumn{17}{c}{\textbf{Temperature=1}} \\
\midrule[0.3pt]
 & Warm-up & 1.64 & 1.23$\times$ & 1.69 & 1.28$\times$ & 1.67 & 1.17$\times$ & 1.61 & 1.22$\times$ & 1.56 & 1.09$\times$ & 1.63 & 1.20$\times$ & 0.47 & 0.50 & 0.37 \\
\midrule[0.3pt]
\multirow{2}{*}{17K} & \textsc{MNRS-Joint} & \textbf{1.72} & \textbf{1.31$\times$} & \textbf{1.80} & \textbf{1.48$\times$} & \textbf{1.82} & \textbf{1.35$\times$} & \textbf{1.67} & \textbf{1.40$\times$} & \textbf{1.60} & \textbf{1.15$\times$} & \textbf{1.72} & \textbf{1.34$\times$} & \textbf{0.43} & \textbf{0.46} & \textbf{0.36} \\
 & Random-Matched & 1.67 & 1.25$\times$ & 1.69 & 1.28$\times$ & 1.63 & 1.14$\times$ & 1.57 & 1.10$\times$ & 1.49 & 0.90$\times$ & 1.61 & 1.13$\times$ & 0.49 & 0.51 & 0.38 \\
\addlinespace[3pt]
\multirow{2}{*}{34K} & \textsc{MNRS-Joint} & \textbf{1.88} & \textbf{1.61$\times$} & \textbf{1.85} & \textbf{1.53$\times$} & \textbf{2.05} & \textbf{1.71$\times$} & \textbf{1.74} & \textbf{1.55$\times$} & \textbf{1.68} & \textbf{1.40$\times$} & \textbf{1.84} & \textbf{1.56$\times$} & \textbf{0.38} & \textbf{0.41} & \textbf{0.34} \\
 & Random-Matched & 1.73 & 1.31$\times$ & 1.78 & 1.42$\times$ & 1.81 & 1.34$\times$ & 1.69 & 1.44$\times$ & 1.65 & 1.26$\times$ & 1.73 & 1.35$\times$ & 0.42 & 0.45 & 0.35 \\
\addlinespace[3pt]
\multirow{2}{*}{68K} & \textsc{MNRS-Joint} & \textbf{1.91} & \textbf{1.69$\times$} & \textbf{1.93} & \textbf{1.69$\times$} & \textbf{2.06} & \textbf{1.72$\times$} & \textbf{1.78} & \textbf{1.78$\times$} & \textbf{1.62} & \textbf{1.21$\times$} & \textbf{1.86} & \textbf{1.62$\times$} & \textbf{0.37} & \textbf{0.39} & \textbf{0.33} \\
 & Random-Matched & 1.83 & 1.54$\times$ & 1.84 & 1.52$\times$ & 1.88 & 1.46$\times$ & 1.75 & 1.64$\times$ & 1.58 & 1.10$\times$ & 1.78 & 1.45$\times$ & 0.39 & 0.42 & 0.34 \\
\bottomrule
\end{tabular}}
\end{minipage}
}]

We compare \textsc{MNRS-Joint} with Random-Matched at continuation
budgets of 17K, 34K, and 68K examples on LLaVA-1.5-7B.
All branches start from the same one-epoch Mixed 68K+68K warm-up
and train for ten additional epochs. At each budget, Random-Matched
samples randomly within each modality using the MM:TXT counts
determined by \textsc{MNRS-Joint}. These counts are 8,821:8,179,
18,173:15,827, and 35,384:32,616, respectively.
We evaluate each checkpoint under greedy tree and stochastic
four-token chain decoding, as in Table~\ref{tab:joint_selection}.

\paragraph{Risk guidance at matched modality counts.}
At every budget, \textsc{MNRS-Joint} improves accepted length and
speedup on all five benchmarks relative to Random-Matched
(Table~\ref{tab:selection_budget}). Its mean total risk is also lower
in both decoding settings. Under greedy tree decoding, the mean
accepted-length advantage is about 0.18 tokens per verification
cycle at all three budgets. Under stochastic chain decoding, the
advantage narrows from 0.11 tokens at 17K to 0.08 at 68K.

\paragraph{Budget dependence.}
Both methods improve in mean accepted length and speedup as the
budget grows. For \textsc{MNRS-Joint}, increasing the budget from
34K to 68K raises mean accepted length from 3.53 to 3.66 under
greedy tree decoding and from 1.84 to 1.86 under stochastic chain
decoding. In both settings, the larger gain occurs between 17K
and 34K. With ten epochs at every budget, larger datasets also
increase total sample exposure.

%% file: appendix/qwen.tex
\twocolumn[{
\begin{minipage}{\textwidth}
\centering
    \includegraphics[width=0.85\textwidth]
        {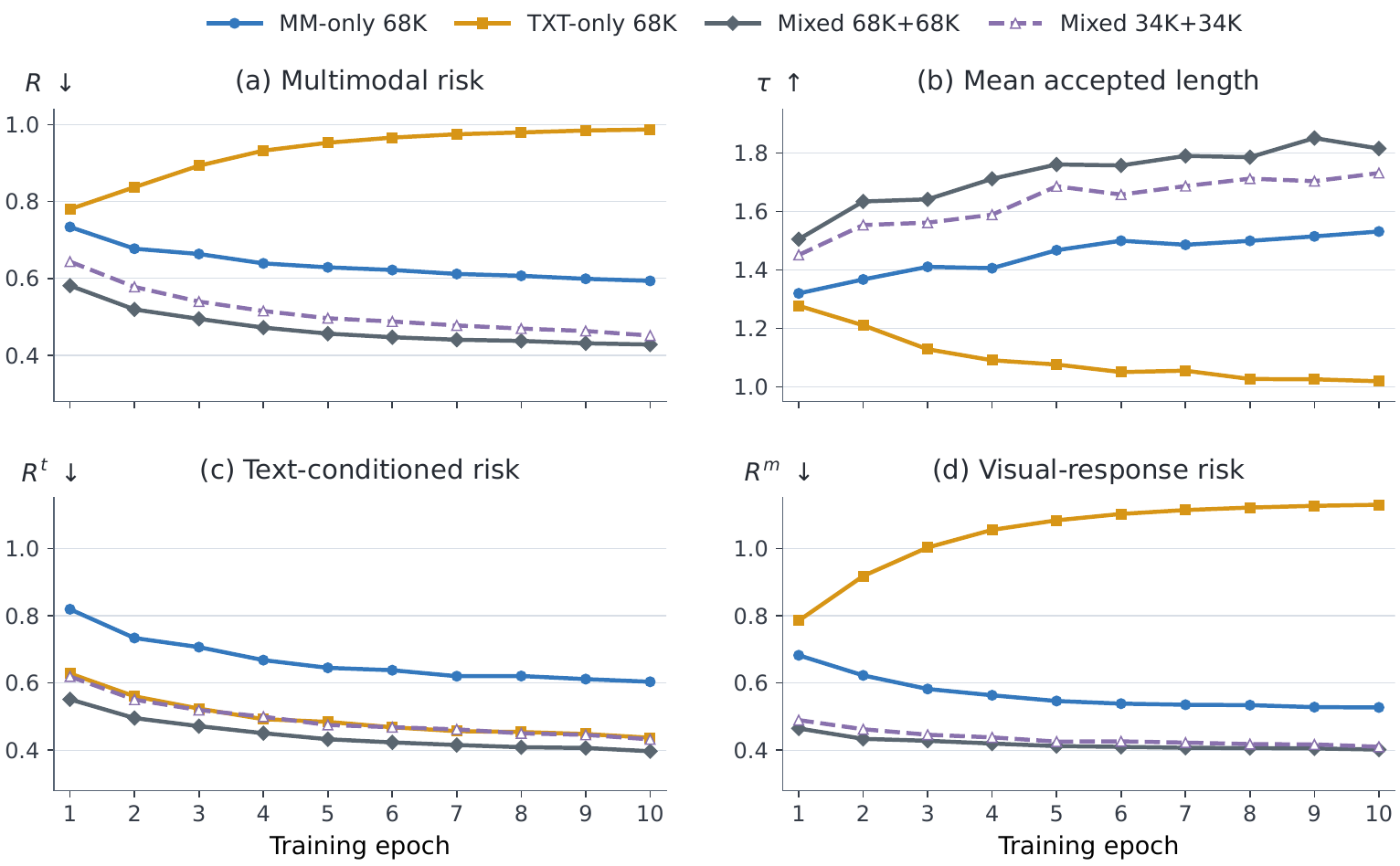}
    \captionof{figure}{
        \textbf{Risk and accepted length under different
        supervision regimes with Qwen2.5-VL-3B under $T=1$ chain decoding.}
        The dashed curve shows the equal-sample-count mixed control.
        Panels (c,d) share a vertical scale.
    }
    \label{fig:qwen_supervision_results}

\end{minipage}
\vspace{6pt}
}]

\section{Additional Results on Qwen2.5-VL-3B}
\label{app:qwen}

We extend the alignment diagnostics and continuation comparisons to
a second target-model family. The experiments examine how supervision
shapes the two alignment components, whether the component risks guide
further training, and how the resulting joint supervision performs.
We evaluate an EAGLE-style draft for Qwen2.5-VL-3B
\citep{qwen25vl}, using the same component-risk definitions
and candidate data sources as in the LLaVA experiments.
Draft dimensions follow the Qwen target architecture.
We evaluate the supervision trajectories under stochastic chain
decoding ($T=1$). The continuation tables compare greedy tree
($T=0$) and stochastic chain ($T=1$) decoding.
Benchmarks, accepted length $\tau$, throughput speedup (Sp.),
and averaging follow Section~\ref{sec:supervision_setup};
risk computation follows Appendix~\ref{app:inference}.
Bold marks the best displayed value within each comparison group,
including ties. The warm-up serves as a reference.

\subsection{Supervision and Component-Wise Alignment}
\label{app:qwen_supervision}

\paragraph{Equal sample counts.} Figure~\ref{fig:qwen_supervision_results}(a,b) compares four
supervision regimes over ten epochs. Both mixed drafts finish
with lower total risk and longer accepted sequences than either
single-modality draft. At the same total sample count as MM-only
training, Mixed 34K+34K reaches a mean accepted length of 1.73,
compared with 1.53 for MM-only.
The equal-count comparison gives both drafts the same total sample
exposure over ten epochs, while changing the supervision composition.
Combining MM and TXT therefore improves alignment and acceptance at
a fixed sample count. Mixed 68K+68K also maintains lower total risk
and higher acceptance than either single-modality regime, extending
the same pattern to the larger mixed dataset.

\paragraph{Component risks.} The clearest contrast appears between TXT-only and Mixed 34K+34K
at epoch ten (Figure~\ref{fig:qwen_supervision_results}(c,d)).
Their text-conditioned risks are close (0.44 and 0.43), yet their
visual-response risks differ sharply (1.13 and 0.41).
This difference accompanies a gap in mean accepted length
(1.02 and 1.73) that text-conditioned risk alone does not reveal.
The paired diagnostic makes this difference visible by separating
agreement on text-only inputs from agreement in the changes induced
by visual input. TXT supervision improves the former while substantial
visual-response disagreement persists. The mixed draft combines
low text-conditioned risk with much lower visual-response risk,
reproducing the alignment pattern observed with LLaVA.

\twocolumn[{
\begin{minipage}{\textwidth}
\centering
\captionof{table}{\textbf{Combined within-modality risk grouping on Qwen2.5-VL-3B.}}
\label{tab:qwen_component_selection}
\footnotesize
\setlength{\tabcolsep}{2.5pt}
\setlength{\aboverulesep}{1.5pt}\setlength{\belowrulesep}{1.5pt}
\renewcommand{\arraystretch}{0.95}
\resizebox{0.88\linewidth}{!}{%
\begin{tabular}{@{}l*{15}{c}@{}}
\toprule
 & \multicolumn{2}{c}{ChartQA} & \multicolumn{2}{c}{MM-Vet} & \multicolumn{2}{c}{ScienceQA} & \multicolumn{2}{c}{TextVQA} & \multicolumn{2}{c}{NoCaps} & \multicolumn{2}{c}{Mean} & \multicolumn{3}{c}{Risk} \\
\cmidrule(lr){2-3}\cmidrule(lr){4-5}\cmidrule(lr){6-7}\cmidrule(lr){8-9}\cmidrule(lr){10-11}\cmidrule(lr){12-13}\cmidrule(lr){14-16}
Method & $\tau$ & Sp. & $\tau$ & Sp. & $\tau$ & Sp. & $\tau$ & Sp. & $\tau$ & Sp. & $\tau$ & Sp. & $R\downarrow$ & $R^t\downarrow$ & $R^m\downarrow$ \\
\midrule
\multicolumn{16}{c}{\textbf{Temperature=0}} \\
\midrule
Warm-up & 2.28 & 1.38$\times$ & 2.20 & 1.45$\times$ & 2.28 & 1.55$\times$ & 1.97 & 1.29$\times$ & 2.26 & 1.35$\times$ & 2.20 & 1.40$\times$ & 0.56 & 0.53 & 0.46 \\
\midrule
\textsc{High} & \textbf{2.36} & \textbf{1.44$\times$} & \textbf{2.36} & \textbf{1.59$\times$} & \textbf{2.50} & \textbf{1.83$\times$} & \textbf{2.11} & \textbf{1.46$\times$} & \textbf{2.47} & \textbf{1.46$\times$} & \textbf{2.36} & \textbf{1.56$\times$} & \textbf{0.51} & \textbf{0.49} & \textbf{0.45} \\
\textsc{Random} & 2.17 & 1.32$\times$ & 2.10 & 1.40$\times$ & 2.16 & 1.37$\times$ & 1.93 & 1.28$\times$ & 2.11 & 1.29$\times$ & 2.09 & 1.33$\times$ & 0.59 & 0.57 & 0.48 \\
\textsc{Low} & 1.96 & 1.19$\times$ & 1.72 & 0.97$\times$ & 1.83 & 1.03$\times$ & 1.64 & 1.04$\times$ & 1.78 & 1.08$\times$ & 1.79 & 1.06$\times$ & 0.69 & 0.69 & 0.52 \\
\midrule
\multicolumn{16}{c}{\textbf{Temperature=1}} \\
\midrule
Warm-up & 1.51 & 1.32$\times$ & 1.49 & 1.27$\times$ & 1.57 & 1.34$\times$ & 1.45 & 1.25$\times$ & 1.51 & 1.24$\times$ & 1.51 & 1.28$\times$ & 0.58 & 0.55 & 0.46 \\
\midrule
\textsc{High} & \textbf{1.55} & \textbf{1.37$\times$} & \textbf{1.50} & \textbf{1.29$\times$} & \textbf{1.57} & \textbf{1.39$\times$} & \textbf{1.45} & \textbf{1.27$\times$} & \textbf{1.54} & \textbf{1.28$\times$} & \textbf{1.52} & \textbf{1.32$\times$} & \textbf{0.56} & \textbf{0.54} & \textbf{0.46} \\
\textsc{Random} & 1.52 & 1.33$\times$ & 1.41 & 1.15$\times$ & 1.44 & 1.17$\times$ & 1.38 & 1.15$\times$ & 1.40 & 1.09$\times$ & 1.43 & 1.18$\times$ & 0.67 & 0.66 & 0.52 \\
\textsc{Low} & 1.44 & 1.20$\times$ & 1.27 & 1.02$\times$ & 1.37 & 1.14$\times$ & 1.29 & 1.11$\times$ & 1.32 & 0.91$\times$ & 1.34 & 1.08$\times$ & 0.71 & 0.71 & 0.53 \\
\bottomrule
\end{tabular}}

\end{minipage}
\vspace{6pt}
}]

\subsection{Training Implications of Component Risks}
\label{app:qwen_selection}

The continuation experiments start from a common warm-up.
We first fix the allocation at 17K MM and 17K TXT examples.
MM groups are ranked by visual-response risk and TXT groups by
text-conditioned risk.
As in Section~\ref{sec:sample_level}, we compare high-risk, randomly
sampled, and low-risk groups. All branches use the same continuation
settings, so the first comparison changes the risk groups while
holding the modality allocation fixed. We then vary the groups in
one modality at a time and compare the two MM ranking signals.

\begin{table}[H]
\centering
\caption{\textbf{Controlled comparisons of component-risk grouping.}}
\label{tab:qwen_single_risk}
\footnotesize
\setlength{\tabcolsep}{2.5pt}
\setlength{\aboverulesep}{1.5pt}\setlength{\belowrulesep}{1.5pt}
\renewcommand{\arraystretch}{1.06}
\begin{tabular}{@{}l*{5}{c}@{}}
\toprule
Method & $\tau$ & Sp. & $R\downarrow$ & $R^t\downarrow$ & $R^m\downarrow$ \\
\midrule
\multicolumn{6}{c}{\textbf{Temperature=0}} \\
\midrule
MM-\textsc{High} & \textbf{2.29} & \textbf{1.50$\times$} & \textbf{0.53} & \textbf{0.50} & \textbf{0.45} \\
MM-\textsc{Random} & 2.09 & 1.33$\times$ & 0.59 & 0.57 & 0.48 \\
MM-\textsc{Low} & 1.98 & 1.26$\times$ & 0.62 & 0.60 & 0.48 \\
\midrule
TXT-\textsc{High} & \textbf{2.26} & \textbf{1.46$\times$} & \textbf{0.54} & \textbf{0.51} & \textbf{0.45} \\
TXT-\textsc{Random} & 2.09 & 1.33$\times$ & 0.59 & 0.57 & 0.48 \\
TXT-\textsc{Low} & 1.96 & 1.23$\times$ & 0.62 & 0.62 & 0.50 \\
\midrule
\multicolumn{6}{c}{\textbf{Temperature=1}} \\
\midrule
MM-\textsc{High} & \textbf{1.47} & \textbf{1.22$\times$} & \textbf{0.61} & \textbf{0.58} & \textbf{0.47} \\
MM-\textsc{Random} & 1.43 & 1.18$\times$ & 0.67 & 0.66 & 0.52 \\
MM-\textsc{Low} & 1.34 & 1.09$\times$ & 0.71 & 0.69 & 0.51 \\
\midrule
TXT-\textsc{High} & \textbf{1.51} & \textbf{1.30$\times$} & \textbf{0.59} & \textbf{0.56} & \textbf{0.47} \\
TXT-\textsc{Random} & 1.43 & 1.18$\times$ & 0.67 & 0.66 & 0.52 \\
TXT-\textsc{Low} & 1.42 & 1.16$\times$ & 0.64 & 0.63 & 0.50 \\
\bottomrule
\end{tabular}
\end{table}

\paragraph{Varying one supervision source.}
Fixing the random subset in one modality reveals the contribution
of the other (Table~\ref{tab:qwen_single_risk}). Under greedy tree
decoding, replacing random MM examples with high-risk MM examples
raises mean accepted length from 2.09 to 2.29. Replacing only the
TXT examples raises it to 2.26. In both decoding settings,
mean accepted length and speedup follow the same
\textsc{High}--\textsc{Random}--\textsc{Low} ordering within each modality.
The MM comparison uses identical TXT examples across branches,
while the TXT comparison uses identical MM examples.
Each component risk thus provides training guidance when the other
supervision source is held fixed. Together with Table~\ref{tab:qwen_component_selection},
these controls show benefits from risk guidance in either source
individually and when both sources are varied together.

\paragraph{Combined within-modality grouping.}
With both modality counts fixed, training on high-risk examples yields
longer accepted sequences and higher speedup than random or low-risk
training on all five benchmarks under both decoding settings
(Table~\ref{tab:qwen_component_selection}). The high-risk branch also
finishes with the lowest mean total risk of the three.
The joint ordering in acceptance and speedup shows that the benefit
of risk-guided continuation carries through to measured decoding
performance. Random and low-risk continuation both finish below the
warm-up in mean accepted length under the two settings. The contrast
with the high-risk branch highlights the importance of which examples
provide the additional supervision.

\begin{table}[H]
\centering
\caption{\textbf{Visual-response versus aggregate risk for MM training.}}
\label{tab:qwen_component_vs_aggregate}
\footnotesize
\setlength{\tabcolsep}{2.5pt}
\setlength{\aboverulesep}{1.5pt}\setlength{\belowrulesep}{1.5pt}
\renewcommand{\arraystretch}{1.06}
\begin{tabular}{@{}l*{5}{c}@{}}
\toprule
MM ranking signal & $\tau$ & Sp. & $R\downarrow$ & $R^t\downarrow$ & $R^m\downarrow$ \\
\midrule
\multicolumn{6}{c}{\textbf{Temperature=0}} \\
\midrule
Aggregate risk $s_i$ & 2.23 & 1.43$\times$ & 0.56 & 0.53 & \textbf{0.45} \\
Visual-response risk $s_i^m$ & \textbf{2.29} & \textbf{1.50$\times$} & \textbf{0.53} & \textbf{0.50} & \textbf{0.45} \\
\midrule
\multicolumn{6}{c}{\textbf{Temperature=1}} \\
\midrule
Aggregate risk $s_i$ & 1.43 & 1.19$\times$ & 0.64 & 0.61 & 0.48 \\
Visual-response risk $s_i^m$ & \textbf{1.47} & \textbf{1.22$\times$} & \textbf{0.61} & \textbf{0.58} & \textbf{0.47} \\
\bottomrule
\end{tabular}
\end{table}

\twocolumn[{
\begin{minipage}{\textwidth}
\centering
\captionof{table}{\textbf{Joint supervision under a shared 34K budget on Qwen2.5-VL-3B.}}
\label{tab:qwen_joint_selection}
\footnotesize
\setlength{\tabcolsep}{2.5pt}
\setlength{\aboverulesep}{1.5pt}\setlength{\belowrulesep}{1.5pt}
\renewcommand{\arraystretch}{0.95}
\resizebox{0.96\linewidth}{!}{%
\begin{tabular}{@{}l*{15}{c}@{}}
\toprule
 & \multicolumn{2}{c}{ChartQA} & \multicolumn{2}{c}{MM-Vet} & \multicolumn{2}{c}{ScienceQA} & \multicolumn{2}{c}{TextVQA} & \multicolumn{2}{c}{NoCaps} & \multicolumn{2}{c}{Mean} & \multicolumn{3}{c}{Risk} \\
\cmidrule(lr){2-3}\cmidrule(lr){4-5}\cmidrule(lr){6-7}\cmidrule(lr){8-9}\cmidrule(lr){10-11}\cmidrule(lr){12-13}\cmidrule(lr){14-16}
Method & $\tau$ & Sp. & $\tau$ & Sp. & $\tau$ & Sp. & $\tau$ & Sp. & $\tau$ & Sp. & $\tau$ & Sp. & $R\downarrow$ & $R^t\downarrow$ & $R^m\downarrow$ \\
\midrule
\multicolumn{16}{c}{\textbf{Temperature=0}} \\
\midrule
Warm-up & 2.28 & 1.38$\times$ & 2.20 & 1.45$\times$ & 2.28 & 1.55$\times$ & 1.97 & 1.29$\times$ & 2.26 & 1.35$\times$ & 2.20 & 1.40$\times$ & 0.56 & 0.53 & 0.46 \\
\midrule
Raw-Joint & 2.26 & 1.37$\times$ & 1.46 & 0.68$\times$ & 2.32 & 1.57$\times$ & 1.24 & 0.54$\times$ & 1.14 & 0.62$\times$ & 1.69 & 0.96$\times$ & 0.73 & \textbf{0.47} & 0.78 \\
Random-Matched & 2.32 & 1.42$\times$ & 2.25 & 1.52$\times$ & 2.38 & 1.66$\times$ & 2.01 & 1.36$\times$ & 2.42 & 1.45$\times$ & 2.28 & 1.48$\times$ & 0.53 & 0.51 & \textbf{0.45} \\
TXT$\rightarrow$MM & 2.05 & 1.27$\times$ & 2.02 & 1.33$\times$ & 2.25 & 1.53$\times$ & 1.82 & 1.18$\times$ & 2.58 & 1.51$\times$ & 2.14 & 1.37$\times$ & 0.58 & 0.54 & 0.48 \\
\textsc{MNRS-Joint} & \textbf{2.38} & \textbf{1.46$\times$} & \textbf{2.37} & \textbf{1.73$\times$} & \textbf{2.51} & \textbf{1.83$\times$} & \textbf{2.15} & \textbf{1.48$\times$} & \textbf{2.61} & \textbf{1.57$\times$} & \textbf{2.40} & \textbf{1.61$\times$} & \textbf{0.50} & 0.48 & \textbf{0.45} \\
\midrule
\multicolumn{16}{c}{\textbf{Temperature=1}} \\
\midrule
Warm-up & 1.51 & 1.32$\times$ & 1.49 & 1.27$\times$ & 1.57 & 1.34$\times$ & 1.45 & 1.25$\times$ & 1.51 & 1.24$\times$ & 1.51 & 1.28$\times$ & 0.58 & 0.55 & 0.46 \\
\midrule
Raw-Joint & 1.68 & \textbf{1.48$\times$} & 1.34 & 1.12$\times$ & 1.55 & 1.31$\times$ & 1.24 & 1.07$\times$ & 1.19 & 0.70$\times$ & 1.40 & 1.13$\times$ & 0.70 & 0.55 & 0.67 \\
Random-Matched & 1.52 & 1.33$\times$ & 1.44 & 1.18$\times$ & 1.56 & 1.31$\times$ & 1.40 & 1.16$\times$ & 1.48 & 1.23$\times$ & 1.48 & 1.24$\times$ & 0.62 & 0.59 & 0.48 \\
TXT$\rightarrow$MM & 1.51 & 1.31$\times$ & 1.48 & 1.26$\times$ & 1.51 & 1.29$\times$ & 1.41 & 1.17$\times$ & 1.57 & 1.29$\times$ & 1.49 & 1.26$\times$ & 0.60 & 0.57 & 0.47 \\
\textsc{MNRS-Joint} & \textbf{1.69} & \textbf{1.48$\times$} & \textbf{1.55} & \textbf{1.39$\times$} & \textbf{1.63} & \textbf{1.41$\times$} & \textbf{1.50} & \textbf{1.29$\times$} & \textbf{1.58} & \textbf{1.35$\times$} & \textbf{1.59} & \textbf{1.39$\times$} & \textbf{0.54} & \textbf{0.52} & \textbf{0.46} \\
\bottomrule
\end{tabular}}

\end{minipage}
\vspace{6pt}
}]

\paragraph{Visual-response versus aggregate risk.}
With the same random TXT subset, ranking MM examples by visual-response
risk yields longer accepted sequences, higher mean speedup, and lower
mean total risk than ranking by aggregate risk under both decoding
settings (Table~\ref{tab:qwen_component_vs_aggregate}).
Under stochastic chain decoding, this change in the MM ranking raises
mean speedup from $1.19\times$ to $1.22\times$.
Because the two branches differ only in their MM ranking signal,
this comparison directly assesses the training value of separating
visual-response disagreement. The advantage on Qwen supports using
that component to guide MM continuation across both target families.

\subsection{Joint Supervision through Relative Risk}
\label{app:qwen_joint}

We next allow the MM:TXT allocation to vary within the 34K budget.
Table~\ref{tab:qwen_joint_selection} uses the same controls as
Section~\ref{sec:joint_evaluation}. Raw-Joint pools raw component
scores. Random-Matched uses the modality counts from \textsc{MNRS-Joint}
and samples randomly within each modality. TXT$\rightarrow$MM uses
the same examples and exposure counts as \textsc{MNRS-Joint},
presenting TXT examples before MM examples.
These controls address three aspects of joint supervision.
Raw-Joint assesses direct pooling of the component scores,
Random-Matched holds the induced composition fixed, and
TXT$\rightarrow$MM holds the selected examples and exposure counts fixed.
The equal-quota high-risk branch in Table~\ref{tab:qwen_component_selection}
provides a further comparison using the same within-modality rankings.

\paragraph{Risk guidance and modality allocation.}
At matched modality counts, \textsc{MNRS-Joint} improves mean accepted
length from 2.28 to 2.40 under greedy tree decoding and from 1.48 to
1.59 under stochastic chain decoding relative to Random-Matched.
Mean speedup also improves in both settings.
Keeping the within-modality rankings and assigning equal MM and TXT
quotas instead yields shorter accepted sequences in both settings
(the \textsc{High} branch in Table~\ref{tab:qwen_component_selection}).
The two controls complement each other. Random-Matched tests the
choice of examples at the MNRS-Joint composition, whereas the
equal-quota branch tests a different composition using the same
risk rankings. MNRS-Joint improves mean acceptance over both.

\paragraph{Joint versus sequential supervision.}
With the examples and their exposure counts held fixed, joint training
yields longer accepted sequences and higher speedup than
TXT$\rightarrow$MM on every benchmark under both decoding settings
(Table~\ref{tab:qwen_joint_selection}).
Under stochastic chain decoding, mean speedup rises from
$1.26\times$ with sequential training to $1.39\times$ with joint training.
The matched exposure counts make the order of supervision the
central difference in this comparison. Interleaving MM and TXT
during continuation improves the final draft across all five
benchmarks, supporting the joint training strategy on Qwen.

\paragraph{Component alignment after joint continuation.}
Raw-Joint has lower text-conditioned risk than MNRS-Joint under
greedy tree decoding, alongside substantially higher visual-response
and total risks. Its mean accepted length also falls below the
warm-up under both decoding settings. This contrast connects the
continuation results to the component diagnosis in
Section~\ref{app:qwen_supervision}. Agreement under the text
condition captures only one part of the draft's multimodal alignment.
MNRS-Joint combines lower total risk with the highest mean accepted
length and speedup among the joint-supervision branches.

\twocolumn[{
\begin{minipage}{\textwidth}
\centering
    \includegraphics[width=0.9\textwidth]{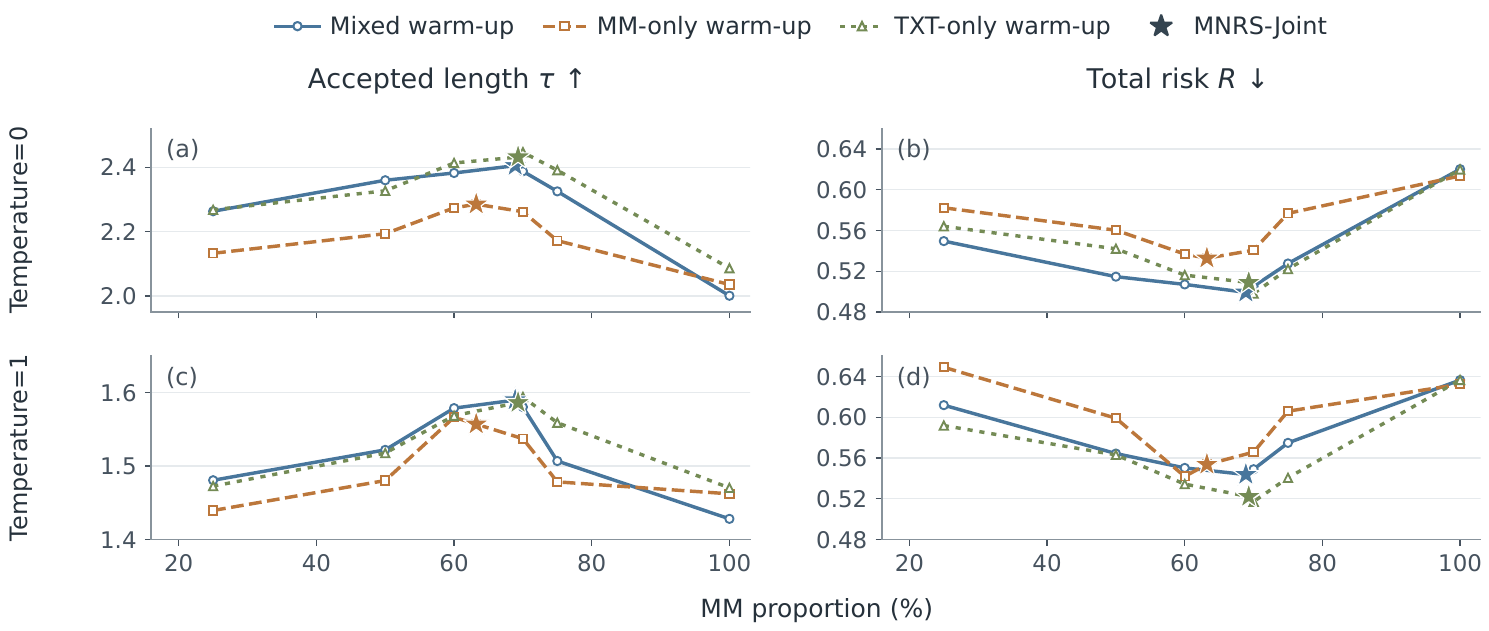}
    \captionof{figure}{\textbf{Modality composition across Qwen warm-ups.}
    Mean accepted length (left) and total risk (right) under greedy tree
    (top) and stochastic chain (bottom) decoding. Stars mark the
    \textsc{MNRS-Joint} MM shares of 68.88\% (Mixed), 63.22\% (MM-only),
    and 69.30\% (TXT-only). Lines connect evaluated mixtures.}
    \label{fig:qwen_ratio_sweep}

\end{minipage}
\vspace{6pt}
}]

\paragraph{Evaluating the supervision mixture.}
We evaluate fixed-budget ratio sweeps from the one-epoch Mixed,
MM-only, and TXT-only warm-ups. Within each regime, we retain the
component-risk rankings and continuation settings and change only
the MM:TXT counts. Both decoding configurations evaluate the
resulting checkpoints. The sweeps locate each precomputed
MNRS-Joint allocation relative to alternative mixtures.

Whereas the favorable MM share shifts across LLaVA warm-ups
(Section~\ref{sec:modality_composition}), the favorable Qwen mixtures
are concentrated around 60--70\% MM
(Figure~\ref{fig:qwen_ratio_sweep}). Before continuation,
\textsc{MNRS-Joint} assigns MM shares of 68.88\%, 63.22\%, and
69.30\% for the Mixed, MM-only, and TXT-only warm-ups, respectively.
These allocations fall in high-performing regions under both
decoding settings.

\textsc{MNRS-Joint} attains the highest observed mean accepted
length with the Mixed warm-up under both decoding settings and
with the MM-only warm-up under greedy tree decoding.
In the other three comparisons, its accepted length remains within
0.61\% of the best evaluated mixture. The corresponding total risks
are also at or near the observed minima.
Moving toward MM-only continuation reduces mean
accepted length and increases total risk in all three regimes
under both decoding settings.
The favorable Qwen mixtures are more similar across warm-ups than
their LLaVA counterparts, while the mean-normalized allocations
remain close to the high-performing region in each Qwen sweep.
Together with the controlled continuation comparisons, these
results support component-risk guidance for choosing examples
and configuring effective joint supervision on a second target family.\par

%% file: appendix/setup.tex
\section{Experimental Details}
\label{app:experimental_details}

\paragraph{Code availability.}
The supplementary material provides anonymized code and
experiment configurations for risk computation, continuation-set
construction, draft training, and benchmark evaluation.
\subsection{Inference and Evaluation Settings}
\label{app:inference}

\paragraph{Decoding configurations.}
For LLaVA-1.5-7B, we use a single-layer EAGLE draft.
Target and draft inference use FP16, batch size one,
and a single GPU.
Greedy tree decoding ($T=0$) uses a dynamic draft tree with
32 nodes including the root, maximum expansion depth 6,
and branching factor 4.
Stochastic chain decoding ($T=1$) proposes up to four draft
tokens per verification cycle, with top-$k$ and top-$p$
filtering disabled. Comparisons between decoding
settings use the same trained checkpoints.

The generation and total sequence-length thresholds are
256 and 4,096 tokens.
Stopping conditions are checked after each verification cycle:
generation ends on a termination token or after reaching
a length threshold, so the final cycle can exceed the
configured threshold.

\paragraph{Evaluation protocol.}
All methods use the same image-containing examples from
ChartQA, MM-Vet, ScienceQA, TextVQA, and NoCaps, with
generation seed 123.
For each benchmark, accepted length $\tau$ is the total
number of generated tokens divided by the number of
target-verification cycles.
The generated-token count includes the additional
target-verified token emitted in each cycle.
Throughput speedup (Sp.) is reported relative to target-only
decoding.
Throughput measurements use a single NVIDIA GeForce RTX 5090 GPU.
The timed interval includes prefill and autoregressive decoding.
Mean accepted length and mean speedup are the unweighted
arithmetic averages of the five benchmark-level values.

\paragraph{Risk computation.}
For both LLaVA and Qwen, we compute full-vocabulary
probabilities from unscaled target and draft logits,
applying softmax and risk calculations in FP32 without
top-$k$ or top-$p$ filtering.
For the paired text condition, we remove the image input
and its marker and recompute target features and draft
states without image-conditioned caches, keeping the
response tokens fixed.
Candidate scores are averaged over assistant-response token
positions in the original training examples, with sequences
truncated to 2,048 tokens.
Benchmark risks use saved responses from a designated reference
model, with a separate response set for each decoding setting.
Within each setting, all evaluated drafts and checkpoints share
the same reference responses and token prefixes. We recompute
target hidden states on these responses and use them as
teacher-forced inputs to the draft. Accepted length and speedup
are measured through actual speculative decoding.
Risks are averaged over response tokens within each example,
then equally over examples within each benchmark and across
the five benchmarks; all components use identical positions
and weights.
Risk definitions follow Section~\ref{sec:risk}.

\subsection{Benchmark-Specific Prompts}
\label{app:benchmark_prompts}

For each target model, we use its chat template and supply the image
followed by the benchmark-specific text as a single user message.
We supply no system message.
We use the test splits of ChartQA and MM-Vet and the validation
splits of ScienceQA, TextVQA, and NoCaps.
For ScienceQA, we evaluate only image-bearing examples.
The templates below specify the task-specific text. All
methods use these templates.

\paragraph{ChartQA.}
We append the following instruction to each benchmark question
to request an explanation:
\begin{quote}
\small\ttfamily
\{question\}\\
Please answer with an explanation.
\end{quote}

\paragraph{MM-Vet.}
We use the original benchmark question as the prompt,
without adding task instructions:
\begin{quote}
\small\ttfamily
\{question\}
\end{quote}

\paragraph{ScienceQA.}
We provide the question and labeled answer choices, requesting
both an answer and a short explanation:
\begin{quote}
\small\ttfamily
You are given an image and a multiple-choice question.\\
Choose the best answer from the choices.

Question: \{question\}\\
Choices:\\
(A) \{choice\_a\}\\
(B) \{choice\_b\}\\
\ldots\\
(\{last\_letter\}) \{last\_choice\}

Please follow this format:\\
Answer: (X) <choice>\\
Reasoning: <short explanation>
\end{quote}
The dataset's reference solution, lecture, and hint are not
included in the input.

\paragraph{TextVQA.}
We append a fixed OCR-oriented instruction to the question:
\begin{quote}
\small\ttfamily
\{question\}\\
Perform an OCR task on the provided image. Extract the text
accurately and provide a detailed explanation of the process.
Ensure the response is comprehensive and well-structured.
\end{quote}
The prompt excludes dataset-provided OCR tokens.

\paragraph{NoCaps.}
We caption all images with the same prompt:
\begin{quote}
\small\ttfamily
Describe the image in one sentence, paying particular attention
to uncommon objects.
\end{quote}

Ground-truth answers and reference captions are excluded from
all prompts.

\subsection{Training Settings}
\label{app:training}

\begin{table}[t]
    \centering
    \caption{\textbf{Draft training hyperparameters.}}
    \label{tab:training_settings}
    \footnotesize
    \setlength{\tabcolsep}{3pt}
    \renewcommand{\arraystretch}{1.08}
    \begin{tabular*}{\columnwidth}{
        @{\extracolsep{\fill}}p{0.43\columnwidth}p{0.53\columnwidth}@{}
    }
        \toprule
        Hyperparameter & Value \\
        \midrule
        Optimizer & AdamW \\
        Peak learning rate & $3\times10^{-5}$ \\
        Adam coefficients & $(0.9,\,0.95)$ \\
        Weight decay & 0 \\
        Learning-rate schedule & Logarithmic warm-up, linear decay \\
        Warm-up steps & 2,000 \\
        Configured schedule horizon & 800,000 optimizer steps \\
        Gradient clipping & 0.5 \\
        Maximum sequence length & 2,048 tokens \\
        Effective batch size & 16 (Section~\ref{sec:supervision_analysis});
            8 (Sections~\ref{sec:sample_level} and~\ref{sec:joint_selection}) \\
        Training precision & BF16 \\
        Hidden-state noise & Uniform, coefficient 0.2 \\
        Gradient checkpointing & Enabled \\
        Random seed & 123 \\
        \bottomrule
    \end{tabular*}
\end{table}
\paragraph{Draft architecture and objective.}
The EAGLE-style draft uses a lightweight single-layer
autoregressive decoder.
The draft takes cached target hidden states together with shifted token
embeddings as input, while the target model, token embedding, and
language-model head remain frozen.
Model-specific hidden dimensions and attention configurations follow the
corresponding target architecture.

We train the draft with a SmoothL1 hidden-state regression loss
and a soft-target cross-entropy loss, weighted by 1.0 and 0.1,
respectively.
The regression loss aligns draft hidden states with target features, while the
cross-entropy term aligns their token distributions through the frozen
language-model head.
Both losses are averaged over supervised response positions
retained after truncation. Loss weights are fixed across experiments.

\paragraph{Optimization.}
Table~\ref{tab:training_settings} lists the optimizer and training
hyperparameters. Gradient accumulation maintains the effective
batch size within each experiment. The configured horizon defines
the learning-rate decay schedule, and each run ends after the
specified number of epochs.

\paragraph{Training schedules and initialization.}
The supervision-regime experiments in
Section~\ref{sec:supervision_analysis} train for ten epochs
from a shared draft initialization.
The joint continuation branches in
Sections~\ref{sec:sample_level} and~\ref{sec:joint_selection}
start from the same one-epoch Mixed warm-up and train
for ten additional epochs.
The ratio sweeps in Section~\ref{sec:modality_composition}
start from the corresponding one-epoch Mixed, MM-only, or
TXT-only warm-up. All ratio branches train for the same
number of continuation epochs.
Within each regime, the 34K budget and within-modality rankings
are fixed across ratios. Each trained checkpoint is evaluated
under both decoding settings.

At branch initialization, we load the warm-up draft weights,
continue the learning-rate schedule from its endpoint,
and reset the AdamW~\citep{loshchilov2019decoupled}
optimizer states.
Resuming an existing branch restores its complete training state.

TXT$\rightarrow$MM uses the same warm-up and examples as
\textsc{MNRS-Joint}, training for ten epochs on TXT followed
by ten epochs on MM.
Joint and sequential continuation therefore give every
selected example ten exposures.
Optimizer and scheduler states are retained when training
switches from TXT to MM.

\paragraph{Cost of risk-guided supervision.}
For LLaVA-1.5-7B, we estimate that scoring a pool of 68K MM and
68K TXT candidates takes 1.09 GPU-hours, based on recorded processing
rates with image-conditioned target features cached.
For comparison, the 1:1 \textsc{High} branch trained on 34K examples
for ten epochs in 2.96 hours on two GPUs (5.93 GPU-hours).
The estimated scoring cost is therefore 18.4\% of this training-loop cost.
Both \textsc{High} and \textsc{MNRS-Joint} reuse the same scores for
a given warm-up and candidate pool.

%% file: main.bbl
\begin{thebibliography}{49}
\providecommand{\natexlab}[1]{#1}
\providecommand{\url}[1]{\texttt{#1}}
\expandafter\ifx\csname urlstyle\endcsname\relax
  \providecommand{\doi}[1]{doi: #1}\else
  \providecommand{\doi}{doi: \begingroup \urlstyle{rm}\Url}\fi

\bibitem[Agrawal et~al.(2019)Agrawal, Desai, Wang, Chen, Jain, Johnson, Batra,
  Parikh, Lee, and Anderson]{NoCaps}
Harsh Agrawal, Karan Desai, Yufei Wang, Xinlei Chen, Rishabh Jain, Mark
  Johnson, Dhruv Batra, Devi Parikh, Stefan Lee, and Peter Anderson.
\newblock nocaps: novel object captioning at scale.
\newblock In \emph{Proceedings of the IEEE International Conference on Computer
  Vision}, pages 8948--8957, 2019.

\bibitem[Alayrac et~al.(2022)Alayrac, Donahue, Luc, Miech, Barr, Hasson, Lenc,
  Mensch, Millican, Reynolds, Ring, Rutherford, Cabi, Han, Gong, Samangooei,
  Monteiro, Menick, Borgeaud, Brock, Nematzadeh, Sharifzadeh, Bi\'{n}kowski,
  Barreira, Vinyals, Zisserman, and Simonyan]{flamingo2022}
Jean-Baptiste Alayrac, Jeff Donahue, Pauline Luc, Antoine Miech, Iain Barr,
  Yana Hasson, Karel Lenc, Arthur Mensch, Katherine Millican, Malcolm Reynolds,
  Roman Ring, Eliza Rutherford, Serkan Cabi, Tengda Han, Zhitao Gong, Sina
  Samangooei, Marianne Monteiro, Jacob~L Menick, Sebastian Borgeaud, Andy
  Brock, Aida Nematzadeh, Sahand Sharifzadeh, Miko\l~aj Bi\'{n}kowski, Ricardo
  Barreira, Oriol Vinyals, Andrew Zisserman, and Kar\'{e}n Simonyan.
\newblock Flamingo: a visual language model for few-shot learning.
\newblock In \emph{Advances in Neural Information Processing Systems}, pages
  23716--23736. Curran Associates, Inc., 2022.

\bibitem[Ankner et~al.(2024)Ankner, Parthasarathy, Nrusimha, Rinard,
  Ragan-Kelley, and Brandon]{hydra}
Zachary Ankner, Rishab Parthasarathy, Aniruddha Nrusimha, Christopher Rinard,
  Jonathan Ragan-Kelley, and William Brandon.
\newblock Hydra: Sequentially-dependent draft heads for medusa decoding, 2024.

\bibitem[Bai et~al.(2025)Bai, Chen, Liu, Wang, Ge, Song, Dang, Wang, Wang,
  Tang, Zhong, Zhu, Yang, Li, Wan, Wang, Ding, Fu, Xu, Ye, Zhang, Xie, Cheng,
  Zhang, Yang, Xu, and Lin]{qwen25vl}
Shuai Bai, Keqin Chen, Xuejing Liu, Jialin Wang, Wenbin Ge, Sibo Song, Kai
  Dang, Peng Wang, Shijie Wang, Jun Tang, Humen Zhong, Yuanzhi Zhu, Mingkun
  Yang, Zhaohai Li, Jianqiang Wan, Pengfei Wang, Wei Ding, Zheren Fu, Yiheng
  Xu, Jiabo Ye, Xi Zhang, Tianbao Xie, Zesen Cheng, Hang Zhang, Zhibo Yang,
  Haiyang Xu, and Junyang Lin.
\newblock Qwen2.5-vl technical report, 2025.

\bibitem[Bhendawade et~al.(2024)Bhendawade, Belousova, Fu, Mason, Rastegari,
  and Najibi]{spec_streaming}
Nikhil Bhendawade, Irina Belousova, Qichen Fu, Henry Mason, Mohammad Rastegari,
  and Mahyar Najibi.
\newblock Speculative streaming: Fast llm inference without auxiliary models,
  2024.

\bibitem[Cai et~al.(2024)Cai, Li, Geng, Peng, Lee, Chen, and Dao]{Medusa}
Tianle Cai, Yuhong Li, Zhengyang Geng, Hongwu Peng, Jason~D. Lee, Deming Chen,
  and Tri Dao.
\newblock Medusa: Simple {LLM} inference acceleration framework with multiple
  decoding heads.
\newblock In \emph{Proceedings of the 41st International Conference on Machine
  Learning}, pages 5209--5235. PMLR, 2024.

\bibitem[Chen et~al.(2023)Chen, Borgeaud, Irving, Lespiau, Sifre, and
  Jumper]{acceleratinglargelanguagemodel}
Charlie Chen, Sebastian Borgeaud, Geoffrey Irving, Jean-Baptiste Lespiau,
  Laurent Sifre, and John Jumper.
\newblock Accelerating large language model decoding with speculative sampling,
  2023.

\bibitem[Chen et~al.(2026)Chen, Liang, and Liu]{DFLASH}
Jian Chen, Yesheng Liang, and Zhijian Liu.
\newblock Dflash: Block diffusion for flash speculative decoding, 2026.

\bibitem[Chen et~al.(2024{\natexlab{a}})Chen, May, Svirschevski, Huang,
  Ryabinin, Jia, and Chen]{sequoia}
Zhuoming Chen, Avner May, Ruslan Svirschevski, Yuhsun Huang, Max Ryabinin,
  Zhihao Jia, and Beidi Chen.
\newblock Sequoia: Scalable and robust speculative decoding.
\newblock In \emph{Advances in Neural Information Processing Systems}, pages
  129531--129563. Curran Associates, Inc., 2024{\natexlab{a}}.

\bibitem[Chen et~al.(2024{\natexlab{b}})Chen, Wu, Wang, Su, Chen, Xing, Zhong,
  Zhang, Zhu, Lu, Li, Luo, Lu, Qiao, and Dai]{internvl}
Zhe Chen, Jiannan Wu, Wenhai Wang, Weijie Su, Guo Chen, Sen Xing, Muyan Zhong,
  Qinglong Zhang, Xizhou Zhu, Lewei Lu, Bin Li, Ping Luo, Tong Lu, Yu Qiao, and
  Jifeng Dai.
\newblock Internvl: Scaling up vision foundation models and aligning for
  generic visual-linguistic tasks.
\newblock In \emph{Proceedings of the IEEE/CVF Conference on Computer Vision
  and Pattern Recognition (CVPR)}, pages 24185--24198, 2024{\natexlab{b}}.

\bibitem[Cheng et~al.(2024)Cheng, Zhang, Zhang, Wang, and Wang]{redrafter}
Yunfei Cheng, Aonan Zhang, Xuanyu Zhang, Chong Wang, and Yi Wang.
\newblock Recurrent drafter for fast speculative decoding in large language
  models, 2024.

\bibitem[Favero et~al.(2024)Favero, Zancato, Trager, Choudhary, Perera,
  Achille, Swaminathan, and Soatto]{favero2024m3id}
Alessandro Favero, Luca Zancato, Matthew Trager, Siddharth Choudhary,
  Pramuditha Perera, Alessandro Achille, Ashwin Swaminathan, and Stefano
  Soatto.
\newblock Multi-modal hallucination control by visual information grounding.
\newblock In \emph{Proceedings of the IEEE/CVF Conference on Computer Vision
  and Pattern Recognition (CVPR)}, pages 14303--14312, 2024.

\bibitem[Fu et~al.(2024)Fu, Bailis, Stoica, and Zhang]{lookahead}
Yichao Fu, Peter Bailis, Ion Stoica, and Hao Zhang.
\newblock Break the sequential dependency of llm inference using lookahead
  decoding, 2024.

\bibitem[Gagrani et~al.(2024)Gagrani, Goel, Jeon, Park, Lee, and
  Lott]{Gagrani2024}
Mukul Gagrani, Raghavv Goel, Wonseok Jeon, Junyoung Park, Mingu Lee, and
  Christopher Lott.
\newblock On speculative decoding for multimodal large language models.
\newblock In \emph{Proceedings of the IEEE/CVF Conference on Computer Vision
  and Pattern Recognition (CVPR) Workshops}, pages 8285--8289, 2024.

\bibitem[He et~al.(2024)He, Zhong, Cai, Lee, and He]{rest}
Zhenyu He, Zexuan Zhong, Tianle Cai, Jason Lee, and Di He.
\newblock {REST}: Retrieval-based speculative decoding.
\newblock In \emph{Proceedings of the 2024 Conference of the {N}orth {A}merican
  Chapter of the {A}ssociation for {C}omputational {L}inguistics: Human
  Language Technologies (Volume 1: Long Papers)}, pages 1582--1595, Mexico
  City, Mexico, 2024. Association for Computational Linguistics.

\bibitem[Hu et~al.(2025)Hu, Guo, Feng, and Zhao]{adaspec}
Yuezhou Hu, Jiaxin Guo, Xinyu Feng, and Tuo Zhao.
\newblock Adaspec: Selective knowledge distillation for efficient speculative
  decoders, 2025.

\bibitem[Kang et~al.(2025)Kang, Shu, Li, Zhai, and Chen]{ViSpec}
Jialiang Kang, Han Shu, Wenshuo Li, Yingjie Zhai, and Xinghao Chen.
\newblock Vispec: Accelerating vision-language models with vision-aware
  speculative decoding.
\newblock In \emph{Advances in Neural Information Processing Systems}, pages
  115511--115532. Curran Associates, Inc., 2025.

\bibitem[Kim et~al.(2024)Kim, Suresh, Papineni, Riley, Kumar, and
  Benton]{draft_refinement}
Taehyeon Kim, Ananda~Theertha Suresh, Kishore Papineni, Michael Riley, Sanjiv
  Kumar, and Adrian Benton.
\newblock Accelerating blockwise parallel language models with draft
  refinement.
\newblock In \emph{Advances in Neural Information Processing Systems}, pages
  34294--34321. Curran Associates, Inc., 2024.

\bibitem[Leng et~al.(2024)Leng, Zhang, Chen, Li, Lu, Miao, and Bing]{vcd}
Sicong Leng, Hang Zhang, Guanzheng Chen, Xin Li, Shijian Lu, Chunyan Miao, and
  Lidong Bing.
\newblock Mitigating object hallucinations in large vision-language models
  through visual contrastive decoding.
\newblock In \emph{Proceedings of the IEEE/CVF Conference on Computer Vision
  and Pattern Recognition (CVPR)}, pages 13872--13882, 2024.

\bibitem[Leviathan et~al.(2023)Leviathan, Kalman, and Matias]{sd-leviathan23a}
Yaniv Leviathan, Matan Kalman, and Yossi Matias.
\newblock Fast inference from transformers via speculative decoding.
\newblock In \emph{Proceedings of the 40th International Conference on Machine
  Learning}, pages 19274--19286. PMLR, 2023.

\bibitem[Li et~al.(2023)Li, Li, Savarese, and Hoi]{blip22023}
Junnan Li, Dongxu Li, Silvio Savarese, and Steven Hoi.
\newblock {BLIP}-2: Bootstrapping language-image pre-training with frozen image
  encoders and large language models.
\newblock In \emph{Proceedings of the 40th International Conference on Machine
  Learning}, pages 19730--19742. PMLR, 2023.

\bibitem[Li et~al.(2024{\natexlab{a}})Li, Wei, Zhang, and Zhang]{EAGLE}
Yuhui Li, Fangyun Wei, Chao Zhang, and Hongyang Zhang.
\newblock {EAGLE}: Speculative sampling requires rethinking feature
  uncertainty.
\newblock In \emph{Proceedings of the 41st International Conference on Machine
  Learning}, pages 28935--28948. PMLR, 2024{\natexlab{a}}.

\bibitem[Li et~al.(2024{\natexlab{b}})Li, Wei, Zhang, and Zhang]{EAGLE2}
Yuhui Li, Fangyun Wei, Chao Zhang, and Hongyang Zhang.
\newblock {EAGLE}-2: Faster inference of language models with dynamic draft
  trees.
\newblock In \emph{Proceedings of the 2024 Conference on Empirical Methods in
  Natural Language Processing}, pages 7421--7432, Miami, Florida, USA,
  2024{\natexlab{b}}. Association for Computational Linguistics.

\bibitem[Li et~al.(2025)Li, Wei, Zhang, and Zhang]{EAGLE3}
Yuhui Li, Fangyun Wei, Chao Zhang, and Hongyang Zhang.
\newblock Eagle-3: Scaling up inference acceleration of large language models
  via training-time test.
\newblock In \emph{Advances in Neural Information Processing Systems}, pages
  136737--136756. Curran Associates, Inc., 2025.

\bibitem[Lin et~al.(2025)Lin, Lin, Zeng, and Ji]{MSD}
Luxi Lin, Zhihang Lin, Zhanpeng Zeng, and Rongrong Ji.
\newblock Speculative decoding reimagined for multimodal large language models,
  2025.

\bibitem[Liu et~al.(2023)Liu, Li, Wu, and Lee]{Visual_instruction_tuning}
Haotian Liu, Chunyuan Li, Qingyang Wu, and Yong~Jae Lee.
\newblock Visual instruction tuning.
\newblock In \emph{Advances in Neural Information Processing Systems}, pages
  34892--34916. Curran Associates, Inc., 2023.

\bibitem[Liu et~al.(2024{\natexlab{a}})Liu, Li, Li, and Lee]{LLaVA}
Haotian Liu, Chunyuan Li, Yuheng Li, and Yong~Jae Lee.
\newblock Improved baselines with visual instruction tuning.
\newblock In \emph{Proceedings of the IEEE/CVF Conference on Computer Vision
  and Pattern Recognition (CVPR)}, pages 26296--26306, 2024{\natexlab{a}}.

\bibitem[Liu et~al.(2025{\natexlab{a}})Liu, Zheng, Muennighoff, Zeng, Dou,
  Pang, Jiang, and Lin]{regmix}
Qian Liu, Xiaosen Zheng, Niklas Muennighoff, Guangtao Zeng, Longxu Dou, Tianyu
  Pang, Jing Jiang, and Min Lin.
\newblock Regmix: Data mixture as regression for language model pre-training.
\newblock In \emph{International Conference on Learning Representations}, pages
  38305--38339, 2025{\natexlab{a}}.

\bibitem[Liu et~al.(2025{\natexlab{b}})Liu, Zheng, and Chen]{liu2024paying}
Shi Liu, Kecheng Zheng, and Wei Chen.
\newblock Paying more attention to image: A training-free method
  for alleviating hallucination in lvlms.
\newblock In \emph{Computer Vision -- ECCV 2024}, pages 125--140, Cham,
  2025{\natexlab{b}}. Springer Nature Switzerland.

\bibitem[Liu et~al.(2024{\natexlab{b}})Liu, Hu, Bailis, Cheung, Deng, Stoica,
  and Zhang]{online_speculative_decoding}
Xiaoxuan Liu, Lanxiang Hu, Peter Bailis, Alvin Cheung, Zhijie Deng, Ion Stoica,
  and Hao Zhang.
\newblock Online speculative decoding.
\newblock In \emph{Proceedings of the 41st International Conference on Machine
  Learning}, pages 31131--31146. PMLR, 2024{\natexlab{b}}.

\bibitem[Loshchilov and Hutter(2019)]{loshchilov2019decoupled}
Ilya Loshchilov and Frank Hutter.
\newblock Decoupled weight decay regularization.
\newblock In \emph{International Conference on Learning Representations}, 2019.

\bibitem[Lu et~al.(2022)Lu, Mishra, Xia, Qiu, Chang, Zhu, Tafjord, Clark, and
  Kalyan]{ScienceQA}
Pan Lu, Swaroop Mishra, Tanglin Xia, Liang Qiu, Kai-Wei Chang, Song-Chun Zhu,
  Oyvind Tafjord, Peter Clark, and Ashwin Kalyan.
\newblock Learn to explain: Multimodal reasoning via thought chains for science
  question answering.
\newblock In \emph{Advances in Neural Information Processing Systems}, pages
  2507--2521. Curran Associates, Inc., 2022.

\bibitem[Masry et~al.(2022)Masry, Long, Tan, Joty, and Hoque]{ChartQA}
Ahmed Masry, Do~Xuan Long, Jia~Qing Tan, Shafiq Joty, and Enamul Hoque.
\newblock {C}hart{QA}: A benchmark for question answering about charts with
  visual and logical reasoning.
\newblock In \emph{Findings of the Association for Computational Linguistics:
  ACL 2022}, pages 2263--2279, Dublin, Ireland, 2022. Association for
  Computational Linguistics.

\bibitem[McKinzie et~al.(2024)McKinzie, Gan, Fauconnier, Dodge, Zhang, Dufter,
  Shah, Du, Peng, Weers, Belyi, Zhang, Singh, Kang, Jain, Hè, Schwarzer,
  Gunter, Kong, Zhang, Wang, Wang, Du, Lei, Wiseman, Yin, Lee, Wang, Pang,
  Grasch, Toshev, and Yang]{MM1}
Brandon McKinzie, Zhe Gan, Jean-Philippe Fauconnier, Sam Dodge, Bowen Zhang,
  Philipp Dufter, Dhruti Shah, Xianzhi Du, Futang Peng, Floris Weers, Anton
  Belyi, Haotian Zhang, Karanjeet Singh, Doug Kang, Ankur Jain, Hongyu Hè, Max
  Schwarzer, Tom Gunter, Xiang Kong, Aonan Zhang, Jianyu Wang, Chong Wang, Nan
  Du, Tao Lei, Sam Wiseman, Guoli Yin, Mark Lee, Zirui Wang, Ruoming Pang,
  Peter Grasch, Alexander Toshev, and Yinfei Yang.
\newblock Mm1: Methods, analysis \& insights from multimodal llm pre-training,
  2024.

\bibitem[Miao et~al.(2024)Miao, Oliaro, Zhang, Cheng, Wang, Zhang, Wong, Zhu,
  Yang, Shi, Shi, Chen, Arfeen, Abhyankar, and Jia]{specinfer}
Xupeng Miao, Gabriele Oliaro, Zhihao Zhang, Xinhao Cheng, Zeyu Wang, Zhengxin
  Zhang, Rae Ying~Yee Wong, Alan Zhu, Lijie Yang, Xiaoxiang Shi, Chunan Shi,
  Zhuoming Chen, Daiyaan Arfeen, Reyna Abhyankar, and Zhihao Jia.
\newblock Specinfer: Accelerating generative large language model serving with
  tree-based speculative inference and verification, 2024.

\bibitem[Samarin et~al.(2026)Samarin, Krutikov, Shevtsov, Skvortsov, Fisin, and
  Golubev]{lk_losses}
Alexander Samarin, Sergei Krutikov, Anton Shevtsov, Sergei Skvortsov, Filipp
  Fisin, and Alexander Golubev.
\newblock Lk losses: Direct acceptance rate optimization for speculative
  decoding, 2026.

\bibitem[Shen et~al.(2026)Shen, Wang, Zhang, Hsieh, Han, Wan, Zhang, Zhang,
  Xiong, Liu, Zhang, Cao, Zhao, and Zhang]{mmspec}
Hui Shen, Xin Wang, Ping Zhang, Yunta Hsieh, Qi Han, Zhongwei Wan, Ziheng
  Zhang, Jingxuan Zhang, Jing Xiong, Ziyuan Liu, Yifan Zhang, Hangrui Cao,
  Chenyang Zhao, and Mi Zhang.
\newblock Mmspec: Benchmarking speculative decoding for vision-language models,
  2026.

\bibitem[Singh et~al.(2019)Singh, Natarajan, Shah, Jiang, Chen, Batra, Parikh,
  and Rohrbach]{TextVQA}
Amanpreet Singh, Vivek Natarajan, Meet Shah, Yu Jiang, Xinlei Chen, Dhruv
  Batra, Devi Parikh, and Marcus Rohrbach.
\newblock Towards vqa models that can read.
\newblock In \emph{Proceedings of the IEEE/CVF Conference on Computer Vision
  and Pattern Recognition (CVPR)}, 2019.

\bibitem[Stern et~al.(2018)Stern, Shazeer, and Uszkoreit]{blockwise}
Mitchell Stern, Noam Shazeer, and Jakob Uszkoreit.
\newblock Blockwise parallel decoding for deep autoregressive models, 2018.

\bibitem[Tong et~al.(2024)Tong, Brown, Wu, Woo, Middepogu, Akula, Yang, Yang,
  Iyer, Pan, Wang, Fergus, LeCun, and Xie]{cambrian1}
Shengbang Tong, Ellis Brown, Penghao Wu, Sanghyun Woo, Manoj Middepogu,
  Sai~Charitha Akula, Jihan Yang, Shusheng Yang, Adithya Iyer, Xichen Pan,
  Ziteng Wang, Rob Fergus, Yann LeCun, and Saining Xie.
\newblock Cambrian-1: A fully open, vision-centric exploration of multimodal
  llms, 2024.

\bibitem[Vo et~al.(2026)Vo, Nguyen, Srey, Tuan, and Nguyen]{tiger}
Quynh Vo, Cong-Duy Nguyen, Ponhvoan Srey, Luu~Anh Tuan, and Thong Nguyen.
\newblock Tiger: Text-conditioned visual gated routing with acceptance
  alignment for multimodal speculative decoding, 2026.

\bibitem[Wang et~al.(2024)Wang, Bai, Tan, Wang, Fan, Bai, Chen, Liu, Wang, Ge,
  Fan, Dang, Du, Ren, Men, Liu, Zhou, Zhou, and Lin]{qwen2vl}
Peng Wang, Shuai Bai, Sinan Tan, Shijie Wang, Zhihao Fan, Jinze Bai, Keqin
  Chen, Xuejing Liu, Jialin Wang, Wenbin Ge, Yang Fan, Kai Dang, Mengfei Du,
  Xuancheng Ren, Rui Men, Dayiheng Liu, Chang Zhou, Jingren Zhou, and Junyang
  Lin.
\newblock Qwen2-vl: Enhancing vision-language model's perception of the world
  at any resolution, 2024.

\bibitem[Wang et~al.(2026)Wang, Li, Du, Zhou, Zhang, and Yang]{SpecFLASH}
Zihua Wang, Ruibo Li, Haozhe Du, Joey~Tianyi Zhou, Yu Zhang, and Xu Yang.
\newblock Specflash: A latent-guided semi-autoregressive speculative decoding
  framework for efficient multimodal generation, 2026.

\bibitem[Xia et~al.(2023)Xia, Ge, Wang, Chen, Wei, and Sui]{SpecDec}
Heming Xia, Tao Ge, Peiyi Wang, Si-Qing Chen, Furu Wei, and Zhifang Sui.
\newblock Speculative decoding: Exploiting speculative execution for
  accelerating seq2seq generation.
\newblock In \emph{Findings of the Association for Computational Linguistics:
  EMNLP 2023}, pages 3909--3925, Singapore, 2023. Association for Computational
  Linguistics.

\bibitem[Xia et~al.(2024)Xia, Yang, Dong, Wang, Li, Ge, Liu, Li, and
  Sui]{sd_survey}
Heming Xia, Zhe Yang, Qingxiu Dong, Peiyi Wang, Yongqi Li, Tao Ge, Tianyu Liu,
  Wenjie Li, and Zhifang Sui.
\newblock Unlocking efficiency in large language model inference: A
  comprehensive survey of speculative decoding.
\newblock In \emph{Findings of the {A}ssociation for {C}omputational
  {L}inguistics: {ACL} 2024}, pages 7655--7671, Bangkok, Thailand, 2024.
  Association for Computational Linguistics.

\bibitem[Xie et~al.(2023)Xie, Pham, Dong, Du, Liu, Lu, Liang, Le, Ma, and
  Yu]{doremi}
Sang~Michael Xie, Hieu Pham, Xuanyi Dong, Nan Du, Hanxiao Liu, Yifeng Lu,
  Percy~S Liang, Quoc~V Le, Tengyu Ma, and Adams~Wei Yu.
\newblock Doremi: Optimizing data mixtures speeds up language model
  pretraining.
\newblock In \emph{Advances in Neural Information Processing Systems}, pages
  69798--69818. Curran Associates, Inc., 2023.

\bibitem[Yu et~al.(2024)Yu, Yang, Li, Wang, Lin, Liu, Wang, and Wang]{MM-Vet}
Weihao Yu, Zhengyuan Yang, Linjie Li, Jianfeng Wang, Kevin Lin, Zicheng Liu,
  Xinchao Wang, and Lijuan Wang.
\newblock {MM}-vet: Evaluating large multimodal models for integrated
  capabilities.
\newblock In \emph{Proceedings of the 41st International Conference on Machine
  Learning}, pages 57730--57754. PMLR, 2024.

\bibitem[Zhang et~al.(2024)Zhang, Gao, Gan, Dufter, Wenzel, Huang, Shah, Du,
  Zhang, Li, Dodge, You, Yang, Timofeev, Xu, Chen, Fauconnier, Lai, You, Wang,
  Dehghan, Grasch, and Yang]{mm15}
Haotian Zhang, Mingfei Gao, Zhe Gan, Philipp Dufter, Nina Wenzel, Forrest
  Huang, Dhruti Shah, Xianzhi Du, Bowen Zhang, Yanghao Li, Sam Dodge, Keen You,
  Zhen Yang, Aleksei Timofeev, Mingze Xu, Hong-You Chen, Jean-Philippe
  Fauconnier, Zhengfeng Lai, Haoxuan You, Zirui Wang, Afshin Dehghan, Peter
  Grasch, and Yinfei Yang.
\newblock Mm1.5: Methods, analysis \& insights from multimodal llm fine-tuning,
  2024.

\bibitem[Zhou et~al.(2024)Zhou, Lyu, Rawat, Menon, Rostamizadeh, Kumar, Kagy,
  and Agarwal]{distillspec}
Yongchao Zhou, Kaifeng Lyu, Ankit~Singh Rawat, Aditya~Krishna Menon, Afshin
  Rostamizadeh, Sanjiv Kumar, Jean-Fran{\c{c}}ois Kagy, and Rishabh Agarwal.
\newblock Distillspec: Improving speculative decoding via knowledge
  distillation.
\newblock In \emph{The Twelfth International Conference on Learning
  Representations}, 2024.

\end{thebibliography}
